\documentclass[letterpaper]{article} %
\usepackage{aaai2026}  %
\usepackage{times}  %
\usepackage{helvet}  %
\usepackage{courier}  %
\usepackage[hyphens]{url}  %
\usepackage{graphicx} %
\usepackage{natbib}  %
\usepackage{caption} %
\usepackage{algorithm}
\usepackage{algorithmic}

\usepackage{newfloat}
\usepackage{listings}
\DeclareCaptionStyle{ruled}{labelfont=normalfont,labelsep=colon,strut=off} %
\floatstyle{ruled}
\newfloat{listing}{tb}{lst}{}
\floatname{listing}{Listing}
\usepackage{amsmath}
\usepackage{amsfonts}
\usepackage{adjustbox}
\usepackage{booktabs}
\usepackage{multirow}
\usepackage{xcolor}
\usepackage{colortbl}

\title{State-Space Hierarchical Compression with Gated Attention and Learnable Sampling for Hour-Long Video Understanding in Large Multimodal Models}
\author{
    Geewook Kim\textsuperscript{\rm 1, 2},
    Minjoon Seo\textsuperscript{\rm 2}
}
\affiliations{

    \textsuperscript{\rm 1}NAVER Cloud AI\\
    \textsuperscript{\rm 2}KAIST AI\\
    gwkim.rsrch@gmail.com, minjoon@kaist.ac.kr
}

\begin{document}

\maketitle

\begin{abstract}
We propose an efficient framework to compress massive video-frame features before feeding them into large multimodal models, thereby mitigating the severe token explosion arising from hour-long videos. Our design leverages a bidirectional state-space model equipped with a gated skip connection and a learnable weighted-average pooling mechanism applied to periodically inserted learned queries. This structure enables hierarchical downsampling across both spatial and temporal dimensions, preserving performance in a cost-effective manner. Across challenging hour-long video understanding tasks, our approach demonstrates competitive results against state-of-the-art models, while significantly reducing overall token budget. Notably, replacing our state-space model with conventional modules results in substantial performance degradation, highlighting the advantages of the proposed state-space modeling for effectively compressing multi-frame video information. Our framework emphasizes resource-conscious efficiency, making it practical for real-world deployments. We validate its scalability and generality across multiple benchmarks, achieving the dual objectives of efficient resource usage and comprehensive video understanding.
\end{abstract}

\begin{links}
    \link{Code}{https://github.com/naver-ai/mambamia}
    \link{Extended version}{https://arxiv.org/abs/2506.13564}
\end{links}

\section{Introduction}
\label{sec:intro}

The ability to process and understand long-form video is rapidly becoming central to the next generation of multimodal AI systems. Large language models (LLMs) augmented with visual input---so-called large multimodal models (LMMs)---have recently achieved impressive results on images and short video clips~\citep{liu2023llava, kim-seo-2024-efficient, li2025llavaonevision}. However, lifting these models to hour-scale video remains a formidable challenge. Simply representing every frame and patch in a long video leads to a dramatic spike in token sequences—often numbering in the hundreds of thousands—far surpassing the capacity of standard models and hardware. Notably, this phenomenon renders state-of-the-art approaches either impractical or inefficient for scalable video understanding~\citep{zhang2025llavamini, zhang2024llavanextvideo}.

A wide range of prior work attempts to mitigate this ``token explosion.'' Per-frame spatial pooling and token pruning can reduce redundancy locally~\citep{li2025llavaonevision, zhang2024llavanextvideo, cheng2024videollama2advancingspatialtemporal, zhang2025videollama3frontiermultimodal}, but they fail to address accumulation over long time horizons. Other approaches depend on task- or query-specific selection~\citep{cheng2025vilamp, shen2025longvu}, trading away flexible context modeling and limiting downstream applicability. Consequently, it remains an open problem to design a general-purpose, learnable system that can efficiently compress both spatial and temporal redundancy, while preserving the wide context needed for robust reasoning about long, real-world videos.

In this paper, we address this limitation with \textbf{MambaMia} (\textbf{Mamba} for \textbf{M}assive \textbf{i}nput-frame \textbf{a}ggregation), a general and modular framework for hierarchical video token compression. Our approach introduces two key innovations: a gated patch aggregation module that integrates spatial and short-range temporal cues via bi-directional state-space (Mamba) modeling, and a lightweight time-axis aggregator that summarizes global video structure and adaptively filters frames using a cumulative, data-driven importance signal. By combining these stages in a hierarchical pipeline, MambaMia systematically distills massive video input to a compact set of tokens that preserves both fine-grained details and holistic temporal context.

What sets our approach apart is not just efficiency, but also its ability to scale to extremely long videos without sacrificing accuracy. On LVBench, a challenging benchmark requiring reasoning over hour-long videos, MambaMia reduces LLM token usage to just 4.7K---yet achieves a score of 44.6, outperforming contemporary models such as mPLUG-Owl3~\citep{ye2025mplugowl} (43.5) and Qwen2-VL~\citep{Qwen2-VL} (42.0). These results demonstrate that our model brings efficient, hour-long video understanding within reach on standard hardware—a key step toward practical, deployable LMMs.

\begin{figure*}[t]
    \begin{center}
        \includegraphics[width=1.0\linewidth]{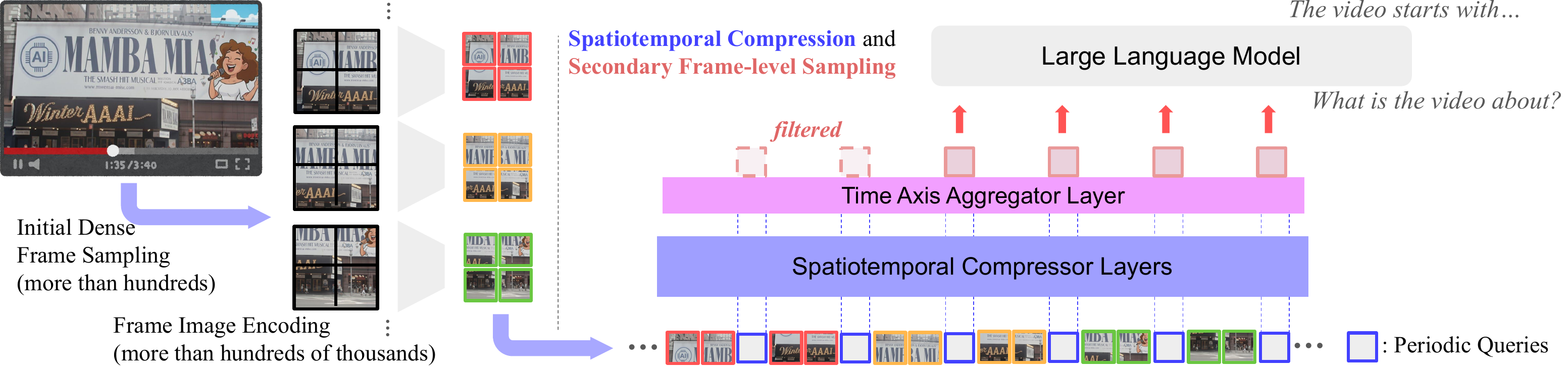}
    \end{center}
    \caption{
Overview of the MambaMia framework.
Given a long video, we densely sample frames and embed patches to form a large sequence of visual tokens.  
Our framework then applies two-stage compression:  
(i) a spatiotemporal compression layer with periodic learnable queries aggregates local features,  
(ii) a time-axis aggregator uses delta-time values for adaptive frame selection.  
This pipeline efficiently reduces token count while preserving rich video context for LLM processing.  
    }
    \label{fig:overall}
\end{figure*}

Our main contributions are as follows.
First, we introduce a unified architecture, MambaMia, which leverages gated patch aggregation (GPA) and a time-axis aggregator (TAA) to deliver highly efficient and accurate token compression for long videos, as demonstrated through extensive and controlled comparisons with prior compression and sequence modeling approaches.
Second, we propose a novel delta-time-based adaptive filtering algorithm, implemented on top of the TAA, which further reduces temporal redundancy by selecting salient frames in a data-driven way; our ablation studies confirm its effectiveness for long-form video inputs.
Finally, we benchmark and release MambaMia alongside the latest large multimodal models, demonstrating that our approach offers state-of-the-art accuracy with drastically fewer LLM tokens and minimal compute, while remaining simple to implement and fast to train from scratch or on top of existing LLM pipelines.

In the following sections, we detail the design, rationale, and empirical benefits of MambaMia, and chart promising future paths for hour-long video understanding research.

\section{Related Work}
\label{sec:related}

\subsubsection{State-Space Models and Mamba}
State-space models (SSMs) have gained traction for handling long sequences efficiently, thanks to their linear complexity compared to the quadratic scaling of attention~\citep{gu2024mamba,NIPS2017_3f5ee243,gpt-neox-20b}. While initially proposed for language modeling, Mamba and related architectures have been adapted for vision tasks as well, demonstrating strong scalability in works such as ViM~\citep{vim}. Recent advances, including bi-directional Mamba variants, have further enhanced video sequence modeling~\citep{videomamba,videomambapark}. Building on these developments, some studies have begun leveraging SSMs for large-scale video feature aggregation and compression in multimodal models. Nonetheless, this area is still emerging, and optimal architectural strategies remain under exploration.

\subsubsection{Spatiotemporal Video Token Compression}
For video understanding in LMMs, it is common practice to treat videos as image sequences, often using per-frame spatial token reduction before temporal concatenation. Various works propose pooling-based spatial compression~\citep{li2025llavaonevision,zhang2024video,zhang2025long,chung2025unifying,cha2023honeybee} and lightweight cross-attention modules like Q-Formers~\citep{zhang2025llavamini,Qwen-VL,Qwen2-VL,Qwen2.5-VL} to reduce token count~\citep{zhang2025llavamini,li2025inference}. However, these 2D reductions, when applied independently per frame, can still lead to very long token sequences for long or densely-sampled videos, resulting in high computational overhead~\citep{zhang2024llavanextvideo,li2025llavaonevision}.

To further mitigate redundancy, some approaches adopt spatiotemporal compression via 3D pooling~\citep{maaz-etal-2024-video,cheng2024videollama2advancingspatialtemporal}, token pruning~\citep{zhang2025videollama3frontiermultimodal}, or attention-based resampling of informative tokens either within or across frames~\citep{li2024llamavid,Li_2024_CVPR_mvbench}. More recently, Mamba-based models have been explored for video token compression, capitalizing on their ability to efficiently aggregate long visual sequences for LLM input. Examples include VAMBA~\citep{vamba}, integrating cross-attention Mamba into LLMs, and BIMBA~\citep{bimba}, which uses periodic non-parametric queries with bi-directional Mamba blocks for summarization. Despite rapid progress, optimal SSM-based video compression architectures for LMMs remain an open research problem.
Additional discussion of related work can be found in Appendix A.

\section{Method}
\label{sec:method}

We address the problem of condensing long-form video into compact, information-rich representations suitable for large language models. As illustrated in Figure~\ref{fig:overall}, our framework comprises two hierarchical stages.

First, a spatiotemporal compression layer aggregates local context across patches and frames into a small set of anchor tokens. Second, a time-axis aggregator models temporal dynamics over these anchors and adaptively selects salient frames. This hierarchical compression enables video understanding over hundreds of frames under a controlled token budget while preserving downstream performance. Below, we describe each component in detail.

\begin{figure*}[t]
    \begin{center}
        \includegraphics[width=0.9\linewidth]{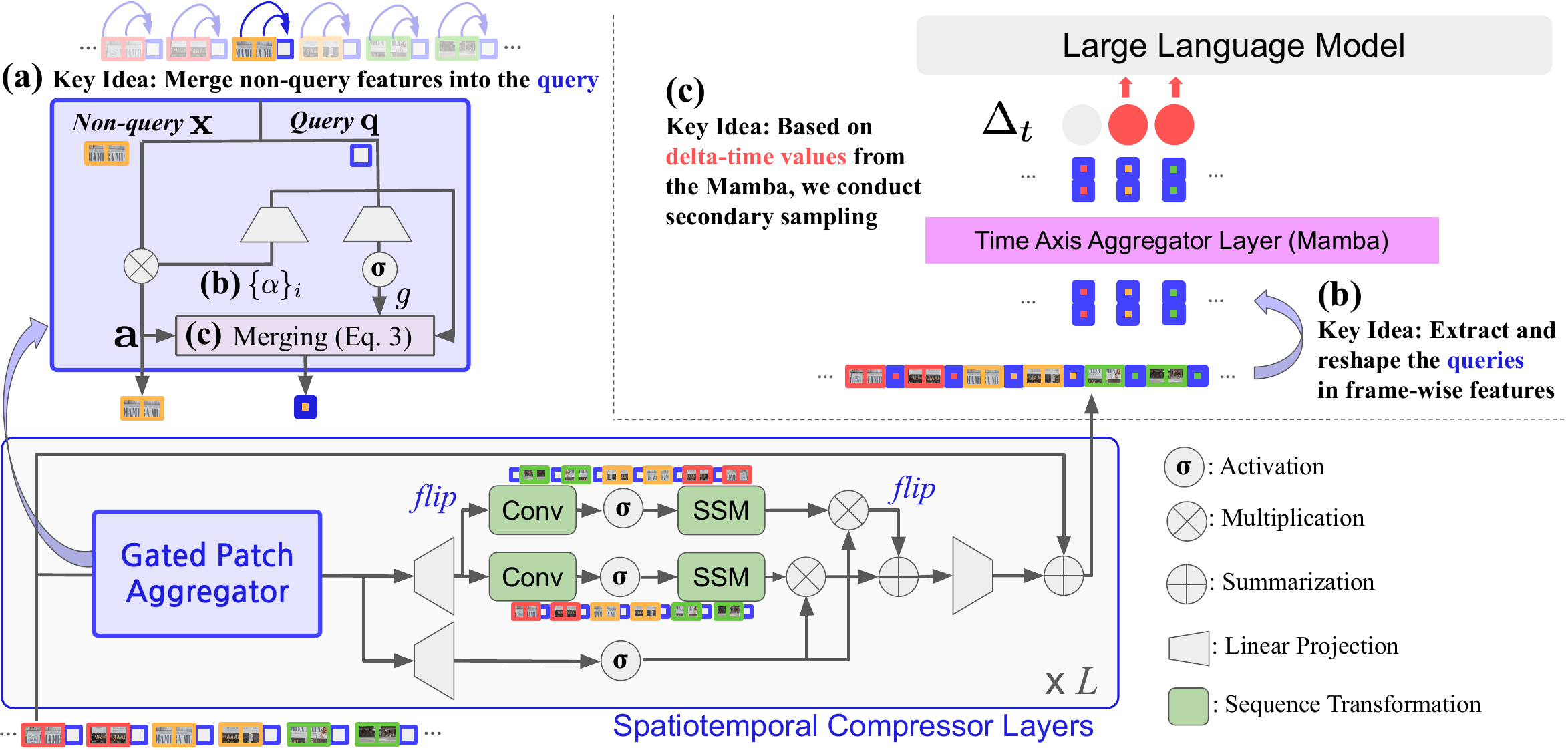}
    \end{center}
\caption{
Architecture.
(a) Periodic query tokens aggregate local context from nearby tokens using learnable pooling and a gating mechanism.  
(b) Frame-wise queries are extracted and reorganized into temporal sequences.  
(c) The time-axis aggregator models temporal dependencies and uses delta-time values for adaptive frame sampling before LLM input.  
}
    \label{fig:overall-b}
\end{figure*}

\subsection{Preliminary: State-Space Models and Mamba}
\label{subsec:preliminary}

A key building block of our approach is the use of SSMs as scalable ``sequence compressors'' between the vision encoder and LLM. SSMs such as the Mamba family~\citep{gu2024mamba,mamba2} enable \emph{linear} computational complexity in sequence length, a critical advantage over the quadratic scaling of Transformers for long-range modeling.

Formally, a discrete SSM updates hidden state $h_t$ with input $x_t$ as:
\begin{equation}
\label{eq:ssm_discrete_main}
h_t = \overline{\mathbf{A}}h_{t-1} + \overline{\mathbf{B}}x_t, \quad y_t=\mathbf{C}h_t,
\end{equation}
where $\overline{\mathbf{A}}, \overline{\mathbf{B}}, \mathbf{C}$ are state-transition and projection matrices (see~\citep{gu2024mamba,mamba2} for details).

In particular, the Mamba architecture achieves high expressivity by dynamically updating select parameters---notably, the input/output mixing and the adaptive step-size $\Delta_t$. The \textbf{adaptive $\Delta_t$} mechanism allows a flexible trade-off between quickly incorporating new input (large $\Delta_t$) and sustaining long-term memory (small $\Delta_t$), \textbf{which is essential for capturing salient events in long sequences}.

We refer readers to~\citep{gu2024mamba,mamba2} and our supplementary material for additional background and derivations. In this work, we leverage these selective SSMs to achieve superior scalability and efficiency for long-form video modeling.

\subsection{Proposed Framework and Architecture}
\label{subsec:mambamia}

Our architecture instantiates the two-stage design above with two main modules:
\begin{itemize}
    \item \textbf{Spatiotemporal Compression Layer with Gated Patch Aggregation (GPA):} Aggregates spatial–temporal patch tokens into a compact set of representative anchor tokens using a bi-directional state-space block and a lightweight gated pooling around learnable query anchors.
    \item \textbf{Time Axis Aggregator (TAA):} Reorganizes anchor tokens by frames, applies a uni-directional state-space block along the temporal axis, and uses its adaptive step size to derive per-frame importance scores for selective frame sampling.
\end{itemize}
Together, GPA and TAA perform hierarchical compression: local aggregation at the patch level followed by adaptive temporal selection. This structure matches the multi-scale nature of real-world videos and provides a favorable trade-off between information retention and computational efficiency. A detailed overview of the pipeline is shown in Figure~\ref{fig:overall-b}; we next describe each module in detail.

\subsubsection{Spatiotemporal Compression Layer with GPA}
\label{subsec:gpa}

Given a video of $M=384$ frames, each resized to $384\times384$ resolution and divided into $16\times16$ image patches, a vision encoder extracts $N=576$ patch embeddings per frame, arranged on a $24\times24$ spatial grid. We then insert learnable query anchors \emph{row-wise}, i.e., one anchor per spatial row that aggregates information from the $k=24$ patches along that row, resulting in a long spatiotemporal token sequence of shape $(230\mathrm{K} \times d_\mathrm{model})$. We set $k=24$ to match the full row width of the $24\times24$ grid.

This sequence is processed by a bi-directional Mamba block~\citep{videomamba,videomambapark} to share information across space and time. To promote explicit and adaptive context aggregation, we introduce the GPA module, wherein each anchor aggregates information from its row-wise neighborhood via a learnable gating function (details in the following and Fig.~\ref{fig:overall-b}(a)). This hybrid module combines the merits of state-space modeling and localized attention, enabling efficient collapse of highly redundant patch-level information into a compressed set of anchor features.

After this layer, we obtain per-frame anchor features, effectively compressing the input sequence while retaining local and global video contexts (e.g., $24$ anchors $\times$ $384$ frames = 9.2K tokens).

\subsubsection{Detail of GPA}
\label{subsec:gated-module}

Figure~\ref{fig:overall-b} (a) illustrates the detailed structure of our proposed GPA module. To explicitly guide information aggregation toward the inserted query tokens, we introduce an adaptive gating mechanism. 
Formally, given a query token \(\mathbf{q} \in \mathbb{R}^{d}\) and its neighboring patch embeddings \(\{\mathbf{x}_i\}_{i=1}^{k}\), we generate aggregation weights \(\{\alpha_i\}\) through a small linear layer (parameters \(\mathbf{W}_\alpha,\mathbf{b}_\alpha\)) followed by a softmax:
\begin{equation}
    \boldsymbol{\alpha} = \mathrm{softmax}\left(\mathbf{W}_\alpha \mathbf{q}+\mathbf{b}_\alpha\right), \quad \mathbf{a}=\sum_{i=1}^{k}\alpha_i \mathbf{x}_i.
\end{equation}
Note that the aggregation weights $\boldsymbol{\alpha}$ are intentionally conditioned only on the query token $\mathbf{q}$. This design implements a lightweight, query-conditioned pooling mechanism, avoiding the computational overhead of full content-aware attention at this fine-grained level.

Next, we compute a scalar gate \(g \in [0,1]\) from the query representation \(\mathbf{q}\) using another linear layer (parameters \(\mathbf{W}_g,\mathbf{b}_g\)) and sigmoid function \(\sigma(\cdot)\):
\begin{equation}\label{eq:final_merge}
g=\sigma(\mathbf{W}_g \mathbf{q} + b_g),\quad \mathbf{f}= (1 - g)\,\mathbf{q}+g\,\mathbf{a}.
\end{equation}

This learnable scalar gate \(g\) adaptively modulates how much neighboring token information replaces the original query representation: \(g\approx0\) preserves previous query contexts, while \(g\approx1\) heavily aggregates local information. Through this adaptive gating, each query token selectively captures key neighboring context, efficiently summarizing both local details and broader spatiotemporal contexts. Collecting these updated query features \(\mathbf{f}\) over all anchors and frames yields the per-frame anchor tokens \(\{f_t\}\) that serve as the input to the TAA.

\subsubsection{Time Axis Aggregator for Adaptive Temporal Summarization}
\label{subsec:taa}

After spatiotemporal compression, the aggregated anchor tokens from all
frames are concatenated into a 1D temporal sequence with shape
$(M,\ 24 \times d_\mathrm{model})$ (e.g., $M = 384$). This sequence is
processed by the Time Axis Aggregator (TAA), a uni-directional Mamba
block along the temporal axis.

As discussed in the preliminary section, Mamba (both v1 and v2)
implements a state-space model with an \emph{adaptive step size}
$\Delta_t$ at each position, which controls how far the internal state
is advanced between tokens (larger $\Delta_t$ induces a larger state
update). We reuse this internal adaptive step size for each frame-level
feature $f_t$~\citep{gu2024mamba,mamba2}, which is computed by a small
scalar head:
\begin{equation}
\label{eq:taa_delta_calc}
\Delta_t = \mathrm{softplus}(\mathbf{W}_\Delta f_t + \mathbf{b}_\Delta),
\end{equation}
yielding a non-negative scalar per frame. We interpret these
$\Delta_t$ values as per-frame importance scores: frames with higher
$\Delta_t$ are treated as more salient for the downstream task. Since
Equation~\eqref{eq:taa_delta_calc} is fully differentiable,
$\mathbf{W}_\Delta$ and $\mathbf{b}_\Delta$ are learned end-to-end via
the main LLM loss, so the model automatically assigns large $\Delta_t$
to informative frames.

\begin{algorithm}[t]
\caption{Cumulative Delta-based Frame Sampling}
\begin{algorithmic}[1]
\REQUIRE Frame-level anchor features $\{f_1, \dots, f_M\}$, importance scores (delta values) $\{\Delta_1, \dots, \Delta_M\}$, delta threshold $\delta_{\text{thresh}}$
\ENSURE Sampled set of key frame features

\STATE Initialize accumulator $A \leftarrow 0$
\STATE Initialize selected set $\mathcal{S} \leftarrow \emptyset$
\FOR{$t = 1$ to $M$}
    \STATE $A \leftarrow A + \Delta_t$
    \IF{$A \geq \delta_{\text{thresh}}$}
        \STATE Add $f_t$ to $\mathcal{S}$
        \STATE Reset $A \leftarrow 0$
    \ENDIF
\ENDFOR
\RETURN $\mathcal{S}$
\end{algorithmic}
\label{algorithm:delta}
\end{algorithm}

To retain only the most informative frames, we apply cumulative
delta-based sampling (Algorithm~\ref{algorithm:delta}): we accumulate
$\Delta_t$ over time and select a frame whenever the sum exceeds a
threshold $\delta_{\text{thresh}}$, then reset the accumulator. This
adaptively reduces the sequence length (e.g., from 384 to 192 frames)
while focusing on frames with large $\Delta_t$. The selected frame
features are then reshaped back to a sequence
(e.g., $(192 \times 24, d_\mathrm{model})$) before being passed to the
LLM. We add a residual connection around the TAA so that it functions
primarily as an importance-based selector, minimally altering the
aggregated features.

Delta-based sampling is especially advantageous for long videos (e.g., 384 frames), while its computational benefits are less pronounced for shorter clips.
Accordingly, we apply delta-based secondary sampling only to our final model and large-scale settings; this length-aware policy applies the secondary sampling only to inputs exceeding $M_{\text{thresh}}$ frames, effectively bypassing it for shorter clips. Comprehensive ablations analyzing its impact are provided in the following sections and in the supplementary material.

\begin{figure*}[t!]
    \begin{center}
        \includegraphics[width=0.97\linewidth]{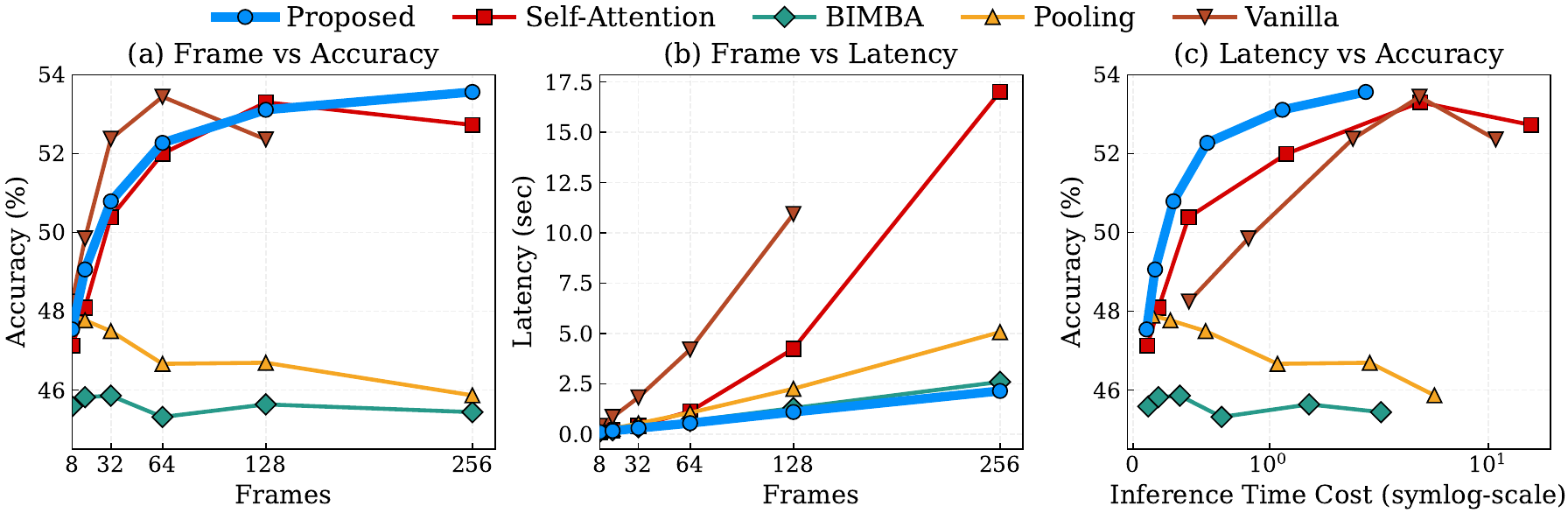}
    \end{center}
\caption{
Trade-off analysis between input frames, inference latency, and accuracy.
We benchmark five models (Proposed, BIMBA, Self-Attention, Pooling, and Vanilla) over varying input frame counts.
(a) Average accuracy across LVBench, MLVU, and VideoMME as a function of frame number;
(b) Inference latency (seconds) versus frame number;
(c) Average accuracy as a function of inference time cost (symlog scale).
The proposed method achieves the best balance, maintaining high accuracy with low test-time compute even as sequence length increases. Notably, (c) shows that our method consistently delivers the highest accuracy under any compute budget, clearly outperforming baselines—especially in high-latency regimes.
}
    \label{fig:aaai26_frame_latency_accuracy}
\end{figure*}

\begin{figure}[t!]
    \begin{center}
        \includegraphics[width=1.0\linewidth]{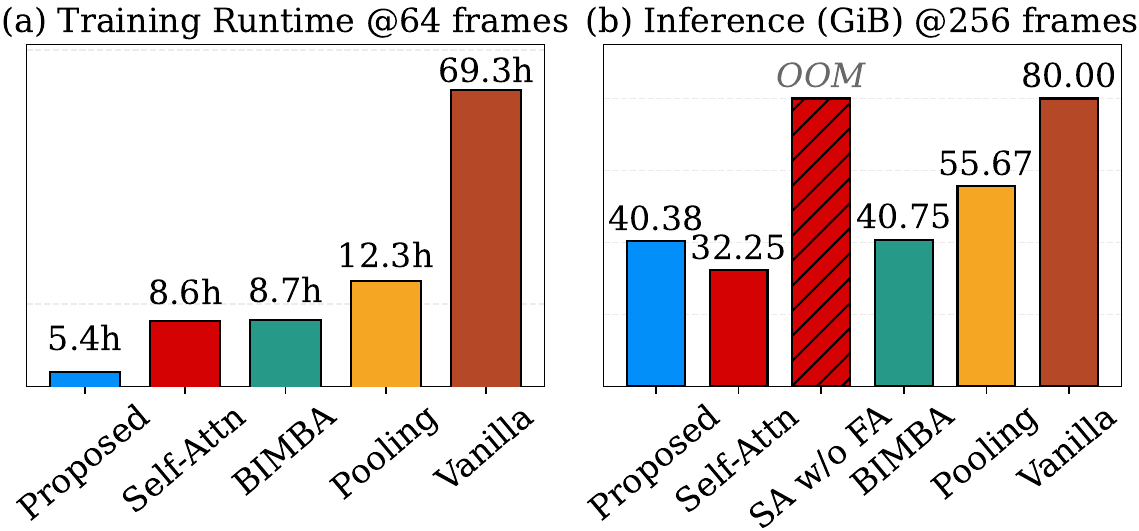}
    \end{center}
    \caption{Comparison of training and inference costs for all baselines. (a) Training runtime (8A100-hours) at 64-frame input. (b) Peak per-GPU memory usage during inference at 256 frames. All models leverage FlashAttention-2~\citep{dao2023flashattention2fasterattentionbetter} where possible. Self-Attention without Flash-Attn (red hatched bar) causes OOM errors at 256 frames.}
    \label{fig:bar_costs}
\end{figure}

\section{Experiments}
\label{sec:exp}

\subsection{Experimental Setup}
\label{subsec:exp_setup}
We summarize the essential elements of our experimental design here. Supplementary materials contain comprehensive information, including all dataset splits, exact hyperparameter settings (such as random seeds), software/hardware environments, and main scripts used in our experiments. 

\subsubsection{Training}
We first construct an image understanding model following the LLaVA-style pipeline~\citep{liu2023llava}, using 1 million instructional images collected according to the Elva recipe~\citep{kim-seo-2024-efficient}. Next, we introduce our compression layers and perform modular alignment on the LLaVA-Pretrain dataset~\citep{liu2023llava}, where only the parameters of the compression layers are updated.
Finally, the language model is unfrozen, and we conduct video-level instruction tuning using both the LLaVA-Video-Set~\citep{zhang2024video} and a subset of the Vista-Set~\citep{ren2024vistaenhancinglongdurationhighresolution}. For ablation studies, we use a subset of 133K video samples, while the final comparative model is trained on the full 1.4 million samples.

\subsubsection{Implementation}
\label{subsubsec:impl_details}
The default setup employs SigLIP2~\citep{tschannen2025siglip} as the vision encoder (producing $N=576$ tokens from $384\times384$ pixels), paired with Qwen2-7B~\citep{yang2024qwen2technicalreport} as the language backbone. The vision encoder remains frozen throughout, consistent with efficient training practices~\citep{liu2023improvedllava,kim-seo-2024-efficient}. 
For some ablation experiments, we also test CLIP-ConvNeXt-Large~\citep{liu2022convnet}, Vicuna-7B~\citep{vicuna2023}, Mamba2-2.7B~\citep{mamba2}, and Pythia-2.8B~\citep{biderman2023pythiasuiteanalyzinglarge}.

Videos are subsampled at up to 2.0 fps, with 64 frames for ablations and 128 frames for final training. At inference, various frame sampling strategies are considered. 
Module-only alignment (LM frozen) uses an initial learning rate of $1\times10^{-4}$, reduced to $2\times10^{-5}$ during full multimodal fine-tuning. 
Our MambaMia block processes queries row-wise ($k=24$) with $(M_{\text{thresh}}, \delta_{\text{thresh}})$ values of $(64, 0.5)$ for Vicuna-based, $(180, 0.6)$ for Qwen-based models (see Appendix J for the sensitivity analysis).
All primary experiments use a fixed random seed (2025) for reproducibility. Furthermore, where computationally practical, we provide additional statistical validation over multiple independent trials in Appendix H.

\subsubsection{Evaluation Benchmarks}
\label{subsubsec:benchmarks}
Our main objective is rigorous assessment of token compression architectures for hour-long video understanding. We focus on LVBench~\citep{wang2024lvbench} and MLVU~\citep{MLVU}, both requiring reasoning over hour-long videos. To contextualize findings, we test five more benchmarks: HourVideo~\citep{chandrasegaran2024hourvideo}, VideoMME (w/o subtitles)~\citep{fu2024video}, LongVideoBench~\citep{longvideobench}, VNBench~\citep{zhao2024videoniah}, and NExTQA~\citep{xiao2021next}.
Further details on the benchmark datasets used are provided in Appendix F.

\begin{table*}[t!]
    \begin{adjustbox}{width=\linewidth}
    \centering
    \begin{tabular}{lcccccccccccc}
        \toprule
        \multirow{2}{*}{\shortstack[l]{\textbf{Model}\\ {Duration}}} 
        & \multirow{2}{*}{\textbf{Year}}
        & \multirow{2}{*}{\shortstack[c]{\textbf{Inst. Data}\\(Img\,/\,Vid)}}
        & \multirow{2}{*}{\textbf{LLM}}
        & \multirow{2}{*}{\shortstack[c]{\textbf{\# Max}\\\textbf{Tokens}$^{\ast}$}}
        & \multirow{2}{*}{\shortstack[c]{\textbf{LVBench}\\30--140min}}
        & \multirow{2}{*}{\shortstack[c]{\textbf{MLVU}\\3--120min}}
        & \multirow{2}{*}{\shortstack[c]{\textbf{HVid}\\$<$95min}}
        & \multirow{2}{*}{\shortstack[c]{\textbf{MME}\\$<$60min}}
        & \multirow{2}{*}{\shortstack[c]{\textbf{LongV}\\$<$60min}}
        & \multirow{2}{*}{\shortstack[c]{\textbf{VNB}\\$<$3min}}
        & \multirow{2}{*}{\shortstack[c]{\textbf{NQA}\\$<$3min}} \\
        & & & & & & & & & & \\
        \midrule
        LongVA & Arxiv24 & 779K (Img only) & Qwen2-7B  & 18K+ & - & 56.3 & -& 52.6 & 51.8 & 41.5 & 68.3 \\
        Qwen2-VL & Arxiv24 & n/a & Qwen2-7B  & - & 42.0 & 64.2 & -& \textbf{63.3} & 55.6 & 33.9 & - \\
        VideoLLaMA2.1 & Arxiv24 & 0.8M / 0.5M & Qwen2-7B  & 1.2K & - & - & -& 54.9 & - & - & - \\
        LLaVA-Video$^{\dagger}$ & Arxiv24 & 4.8M / 1.7M & Qwen2-7B & 6.3K & 40.3 & \underline{66.3} & 33.5& \underline{62.4} & \textbf{58.0} & 55.9 & \textbf{81.8} \\
        LLaVA-OneVision$^{\dagger}$ & TMLR25 & 8.4M / 0.4M & Qwen2-7B & 6.3K & 40.8 & 65.2 & 33.6& 58.5 & 56.6 & 40.4 & 79.3 \\
        mPLUG-Owl3 & ICLR25 & n/a & Qwen2-7B  & - & 43.5 & - & -& 53.5 & 52.1 & - & 78.6 \\
        Video-XL & CVPR25 & 762K / 100K & Qwen2-7B  & - & 36.8 & 64.9 & -& 55.5 & 49.5 & \textbf{61.6} & 77.5 \\
        LongVU & ICML25 & 3.2M / 553K & Qwen2-7B & 8K & 37.8 & 65.4 & -& 55.3 & 53.5 & - & 78.0 \\ 
        VAMBA & ICCV25 & 4.8M / 1.7M & Qwen2-7B  & - & \underline{42.1} & 65.9 & -& 57.8 & 55.9 & - & 78.1 \\
        \rowcolor{green!20}
        \textbf{Proposed} & & 1M / 1.4M & Qwen2-7B & 4.7K & \textbf{44.6} & \textbf{68.0} & \textbf{39.9}& 58.3 & \underline{57.1} & \underline{56.5} & \underline{80.6} \\
        \midrule
        Video-LLaVA & EMNLP24 & 665K / 100K & Vicuna-7B & 2K & - & \underline{47.3} & -& 39.9 & 39.1 & - & - \\ 
        LLaMA-VID & ECCV24 & 625K / 107K & Vicuna-7B & 16K+ & - & 33.2 & -& 25.9 & - & 10.8 & - \\ 
        LLaVA-NeXT-Video$^{\dagger}$ & Blog24 & 760K / 100K & Vicuna-7B & 2.3K & \underline{30.3} & 36.9 & 27.5& 34.1 & \underline{44.1} & \underline{17.3} & 53.6 \\ 
        BIMBA-LLaVA & CVPR25 & 370K (Vid only) & Vicuna-7B  & 2.3K & - & 47.2 & -& \underline{45.7} & - & - & \underline{72.4} \\
        \rowcolor{green!20}
        \textbf{Proposed}$_{\text{Mini}}$ & & 133K (Vid only) & Vicuna-7B & 2.3K & \textbf{37.9} & \textbf{58.5} & \textbf{33.7}& \textbf{47.8} & \textbf{48.3} & \textbf{24.2} & \textbf{73.3} \\
        \bottomrule
    \end{tabular}
    \end{adjustbox}
\caption{
Comparison with Modern LMMs.
We group models by their backbone LLM and mainly include recent, representative models with reproducible and from-scratch training pipelines. 
The table summarizes each method's publication venue, scale of training data for both images and videos, backbone LLM, and the maximum number of tokens processed for evaluation (measured on LVBench). Performance across diverse long video benchmarks is reported (higher is better).
$^{\ast}$ Maximum tokens are based on best available information from original papers or released code.
$^{\dagger}$ Results marked are reproduced under identical evaluation settings for fair comparison; for details, see supplementary material in the extended version.
}
    \label{tab:vgtable}
\end{table*}

\subsubsection{Comparison Methods and Baselines}
\label{subsubsec:baselines}

Our experimental analyses aim to (1) identify the most effective methods for compressing video features, and (2) quantify their performance relative to state-of-the-art models. Experiments are structured accordingly.

We first compare five compression approaches under controlled conditions:  
(1) self-attention-based compression~\citep{Qwen-VL};  
(2) per-frame 2D average pooling~\citep{zhang2024video,li2025llavaonevision};  
(3) a vanilla no-compression baseline (upper bound);  
(4) the BIMBA~\citep{bimba} architecture (Mamba-based baseline); and  
(5) our proposed module. All other factors are held constant, isolating the effect of compression.

We then benchmark against leading LMMs, including LongVA~\citep{zhang2025long}, Qwen2-VL~\citep{Qwen2-VL}, Video-XL~\citep{videovlcvpr}, LongVU~\citep{shen2025longvu}, VAMBA~\citep{vamba}, Video-LLAVA~\citep{lin-etal-2024-video}, LLAMA-VID~\citep{li2024llamavid}, LLAVA-NeXT-Video~\citep{zhang2024llavanextvideo}, BIMBA-LLAVA~\citep{bimba}, LLaVA-Video~\citep{zhang2024video}, and LLaVA-OneVision~\citep{li2025llavaonevision}.  
Ablation configurations and more experimental specifics are detailed in the supplementary materials.

\subsection{Controlled Comparisons of Compression Layers}
\label{subsec:comparsion_in_unified}

To isolate the effect of different video compression and sequence modeling strategies, we conduct controlled experiments where only the compression layer is varied, with all other factors (e.g., data and LLM) held constant. This ensures a fair comparison of efficiency and scalability.

As shown in Figure~\ref{fig:aaai26_frame_latency_accuracy}, our method consistently achieves the best balance of accuracy, inference speed, and token efficiency across varying input lengths. Accuracy remains high even as input frames increase, without the steep latency or memory costs observed in self-attention and pooling baselines. Notably, our approach maintains practical inference time and resource use, while methods such as self-attention and vanilla become increasingly intractable for longer videos.

Figure~\ref{fig:bar_costs} further highlights our method's lower computational and memory footprint during \textbf{both training and inference}, due to effective compression. In contrast, methods lacking a dedicated compression layer face significant inefficiency or out-of-memory errors at scale.

These results demonstrate that the proposed framework enables scalable long-video processing with a modest compute budget, driven by advances in compression and sequence modeling rather than increased compute or data.

\subsection{Benchmark Comparison to Modern LMMs}
\label{subsec:sota_comparison}

We benchmark our proposed models against recent LMMs on a diverse suite of long-video understanding tasks (see Table~\ref{tab:vgtable}). Our approach, based on Qwen2-7B, achieves competitive or state-of-the-art scores across key benchmarks—most notably, reaching 44.6 on LVBench and 68.0 on MLVU—while maintaining a substantially lower token budget (4.7K) compared to competing methods. Furthermore, our strong performance on VNBench, a \textit{needle-in-a-video-haystack} benchmark, addresses concerns on missing brief yet critical events. This demonstrates that our architecture delivers both accuracy and efficiency, even as other models rely on larger compute and token counts.

To ensure fair comparison with Vicuna-based models, we additionally train a lightweight variant (\textbf{Proposed}$_{\text{Mini}}$) under similar data and training conditions. This mini model continues to outperform all Vicuna-based LMMs—even with limited training samples and visual tokens—highlighting the scalability and generalizability of our approach.

We emphasize that all our models are trained and evaluated on academic-scale, publicly available datasets with transparent pipelines, enabling reproducible comparison. 
We strictly adhere to matched-setting comparisons; results reported under different conditions (e.g., using proprietary backbones or further fine-tuning, like some models in STORM~\citep{jiang2025tokenefficientlongvideounderstanding} and BIMBA~\citep{bimba}) are discussed accordingly in Appendix A.
See also Fig.~\ref{fig:bar_costs} for efficiency analyses.
While many recent LMMs pursue brute-force scaling of data and tokens, our results show that efficient architectural design and principled compression are key to scalable, accessible long-video understanding.

\section{Analyses and Discussions}
\label{sec:analysis}

\subsubsection{Proposed Module Ablations}

The proposed MambaMia architecture comprises several key components: GPA and TAA. To understand the individual contributions of these modules, we perform a series of ablation studies. As noted, these ablations are conducted using a lighter training recipe for efficiency.
BIMBA, which can be regarded as the base model for MambaMia, utilizes a bi-directional Mamba block and periodically summarizes information into dedicated queries, but it generates these queries through 3D average pooling. We re-implement this baseline and systematically turn the proposed modules on or off to assess their effects.

Table~\ref{tab:module_ablation} presents the ablation results.  
First, replacing the average pooling query with our learnable GPA leads to marked improvements in performance across all benchmarks.  
While the addition of the TAA does not yield substantial accuracy gains on its own, its value lies in enabling our delta-based frame sampling approach, as discussed in the next section.  
Lastly, we verify that our proposed framework is compatible with both versions of the Mamba block (V1 and V2), with consistently strong results observed in all cases.
We also confirm that the bi-directional spatiotemporal compressor blocks are beneficial; replacing it with uni-directional blocks on our best setting (Table~\ref{tab:module_ablation}, row 3) degrades the average score from 53.56 to 51.86.
As the GPA module can be interpreted as a form of attention, we provide comprehensive qualitative visualizations for it in Appendix K.

\begin{table}[t]
\centering
\begin{adjustbox}{width=\linewidth}
\begin{tabular}{lccccccc}
\toprule
GPA & TAA & Mamba & LVBench & MLVU & MME & Avg. \\
\midrule
             &            & V2 & 35.31 & 53.75 & 47.26 & 45.44       \\
\checkmark   &            & V2 & 41.12 & 62.36 & 53.19 & 52.22       \\
\rowcolor{green!20}
\checkmark   & \checkmark & V2 & 41.06 & 63.96 & 55.67 & 53.56 \\
\checkmark   & \checkmark & V1 & 41.51 & 63.02 & 54.85 & 53.13       \\
\bottomrule
\end{tabular}
\end{adjustbox}
\caption{
Ablation study results for our compression layer architecture design. GPA and TAA denote each module’s usage. The V2 and V1 indicate the Mamba block version.}
\label{tab:module_ablation}
\end{table}

\subsubsection{Delta-based Frame Filtering Ablations}

The TAA layer outputs per-frame delta-time values, which we leverage for an adaptive secondary frame filtering step during inference. This mechanism, incorporated in our final model, further reduces the computational burden (see Table~\ref{tab:vgtable}). For example, with 384 input frames, the initial sequence contains up to 221K patch tokens, which is compressed to about 9K tokens after the first stage ($k=24$ anchors per frame). While this reduction is substantial, directly processing 9K tokens can still be challenging; thus, we apply delta-based frame filtering to select a more compact, information-rich subset. As a result, as shown in Table~\ref{tab:vgtable}, the token budget can be reduced by roughly half, down to 4.7K.

\begin{figure}[t!]
     \centering
     \includegraphics[width=\linewidth]{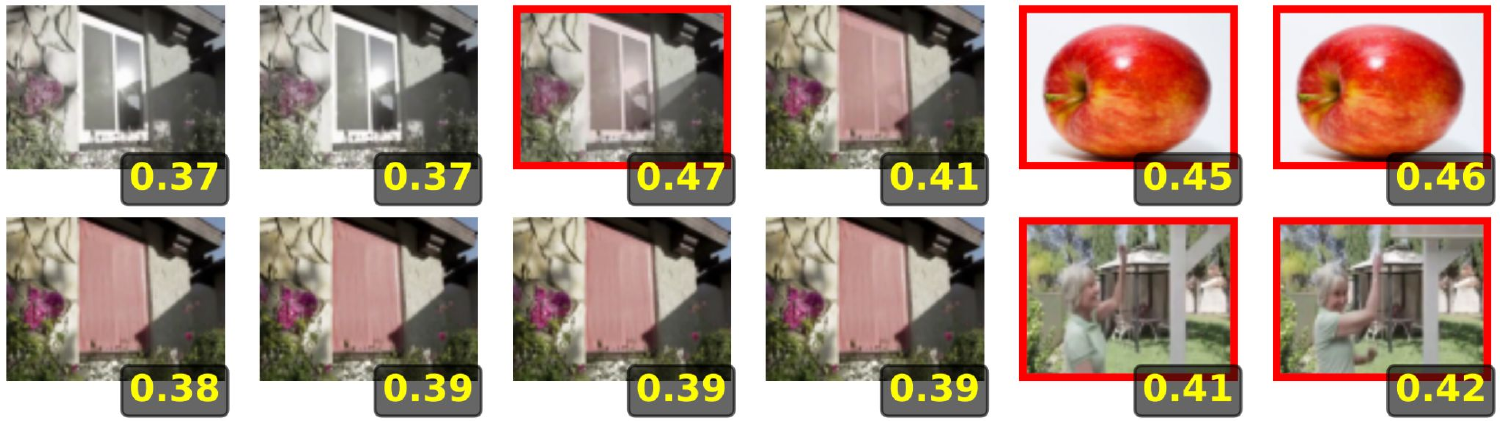}
     \caption{
     Visualization of per-frame delta-time values produced by the TAA layer. Peaks correspond to scene transitions or distinctive events (e.g., \textit{needle-in-a-video-haystack}).
     }
     \label{fig:aaai26_delta_example}
\end{figure}

We compare delta-based filtering (Algorithm~\ref{algorithm:delta})—which retains frames with the highest delta values, targeting approximately 50\% retention using $\delta_{\text{thresh}} = 0.6$—to simple uniform downsampling at the same rate. As shown in Table~\ref{tab:ablation_delta}, delta-based filtering consistently outperforms uniform sampling, demonstrating that the learned delta values are effective for identifying salient frames. To further illustrate this, qualitative analysis in Figure~\ref{fig:aaai26_delta_example} shows that delta spikes often align with scene boundaries or salient events, corresponding to moments of semantic importance.
We provide additional visualizations of per-frame delta-time values, including on diverse video samples, in Appendix K.

\begin{table}[t!]
\centering
\begin{adjustbox}{width=\linewidth}
\begin{tabular}{lccccc}
\toprule
Method  & Sampling Rate & LVBench & MLVU  & MME   & Avg. \\
\midrule
No Sampling      & -  & \textbf{45.58}   & \underline{67.64} & \textbf{58.59} & \textbf{57.27} \\
\cmidrule(lr){1-6}
Uniform & 50\%   & 43.38   & 67.52 & 58.00 & 56.30 \\
\rowcolor{green!20}
Delta-based   & approx. 50\%   & \underline{44.61}   & \textbf{67.99} & \underline{58.26} & \underline{56.95} \\
\bottomrule
\end{tabular}
\end{adjustbox}
\caption{
Ablation of secondary frame sampling methods. Results are shown for no sampling (i.e., disabling this secondary step), uniform 50\% sampling, and delta-based filtering (using a threshold selected to retain half the frames). Delta-based sampling preserves accuracy while halving the input token count.
}
\label{tab:ablation_delta}
\end{table}

\subsubsection{Ablations with Mamba as LLM}
One might wonder whether simply adopting a Mamba-based LLM backbone suffices for efficient long-video modeling, given SSMs’ strong long-range processing capabilities. To address this, we conduct controlled experiments using CLIP-ConvNeXt-Large as the vision encoder, paired with either Pythia-2.8B or Mamba2-2.7B as the language model. As shown in Table~\ref{tab:architecture_llm_comparison}, our compression-augmented method significantly outperforms the vanilla Mamba LLM setting (which uses an MLP projector, per LLaVA~\citep{liu2023improvedllava}), both in throughput and overall accuracy. Similar trends are observed with Pythia. While Mamba LLMs are promising, these results confirm that dedicated compression is crucial for practical long-video processing.

\begin{table}[t!]
\centering
\begin{adjustbox}{width=\linewidth}
\begin{tabular}{llccccc}
\toprule
Method & LLM & Throughput & LVBench & MLVU & MME & Avg. \\
\midrule
Vanilla & Pythia-2.8B & 1.37 & 31.25 & 29.34 & 37.74 & 32.78 \\
Vanilla & Mamba2-2.7B & 1.16 & 30.28 & 40.33 & 31.81 & 34.14 \\
\rowcolor{green!20}
Proposed & Pythia-2.8B & 3.79 & 33.25 & 51.03 & 40.85 & 41.71 \\
Proposed & Mamba2-2.7B & 2.90 & 32.60 & 47.70 & 39.26 & 39.85 \\
\bottomrule
\end{tabular}
\end{adjustbox}
\caption{
Comparison of vanilla and proposed architectures with Pythia-2.8B and Mamba2-2.7B as language models. Compression-augmented models consistently yield higher efficiency and accuracy.
}
\label{tab:architecture_llm_comparison}
\end{table}

\section{Conclusion}

We introduce MambaMia, a unified framework for practical and efficient long-video understanding that integrates novel modular compression with advanced sequence modeling. Through extensive experiments on challenging hour-long video benchmarks, MambaMia demonstrates state-of-the-art accuracy and substantial efficiency gains while operating under realistic memory and compute constraints. Comprehensive ablations and analyses validate the critical importance of effective compression, and adaptive sequence processing across diverse backbones and benchmarks. 
Our codebase and models are publicly available at GitHub to enable reproducibility and further accelerate progress toward scalable, real-world video understanding.

\bibliography{aaai2026,main,anthology}

\appendix
\setcounter{secnumdepth}{1}
\renewcommand{\thesection}{\Alph{section}}
\setcounter{section}{0}
\setcounter{table}{0}
\setcounter{figure}{0}
\renewcommand{\thetable}{\Alph{table}}
\renewcommand{\thefigure}{\Alph{figure}}

\section*{Table of Contents}

\begin{itemize}
    \item \textbf{Section A}: Additional Discussion of Related Work and Positioning of Our Work
    \item \textbf{Section B}: Model Implementation Details
    \item \textbf{Section C}: Computing Infrastructure
    \item \textbf{Section D}: Training Data and Preprocessing
    \item \textbf{Section E}: Full Hyperparameter Listings and Scripts
    \item \textbf{Section F}: Evaluation Benchmarks
    \item \textbf{Section G}: Sanity Checks and Baseline Implementations
    \item \textbf{Section H}: Randomness Analyses and Statistical Tests
    \item \textbf{Section I}: Inference and Training Cost Analysis
    \item \textbf{Section J}: Further Limit Testing
    \item \textbf{Section K}: Additional Visualizations and Examples
    \item \textbf{Section L}: Code and Resource Availability
    \item \textbf{Section M}: Background on Classical SSMs
\end{itemize}

\section{Additional Discussion of Related Work and Positioning of Our Work}
\label{sup:related}

\subsection{Conventional Spatial and Spatiotemporal Aggregation}
Most conventional efforts in video feature compression for large multimodal models (LMMs) adopt a straightforward per-frame spatial token reduction. Various studies have proposed spatial pooling-based compression~\citep{li2025llavaonevision,zhang2024video,zhang2025long,chung2025unifying,cha2023honeybee} and lightweight cross-attention modules~\citep{zhang2025llavamini,Qwen-VL,Qwen2-VL,Qwen2.5-VL} to reduce token count~\citep{zhang2025llavamini,li2025inference}. However, such frame-wise 2D compression schemes face clear limitations when deployed for hour-long videos, where the token budget can increase by orders of magnitude at test time. To address this, several recent works have explored spatiotemporal resampling attention~\citep{Li_2024_CVPR_mvbench,zhang-etal-2023-video,alayrac2022flamingo}, dynamically aggregating features and enabling moderate token budgets without major performance loss for shorter or mid-length clips. Nevertheless, these approaches also exhibit significant performance degradation as input durations or frame densities increase---highlighting the open nature of the problem and motivating continued research into more scalable solutions.

\subsection{Video Token Pruning and Query-aware Compression}
A separate line of research has focused on heuristic or similarity-based pruning~\citep{zhang2025videollama3frontiermultimodal}, which efficiently eliminates redundant spatiotemporal tokens, thereby reducing computation and latency. While lightweight, such approaches risk losing subtle but important details necessary for fine-grained downstream tasks. User-query-aware reduction~\citep{li2024llamavid,shen2025longvu,bimba} aggressively removes vision tokens unlikely to be relevant for a given query, which could yield performance gains when queries are known or can be reliably anticipated. 
Similarly, ViLAMP~\citep{cheng2025vilamp} introduces a query-aware \textit{differential distillation} principle, using query relevance to perform both keyframe selection and patch-level feature merging.
However, the dependence on explicit query availability or accurate prediction significantly limits generality, especially for multi-turn dialogues or genuinely open-ended scenarios where relevant queries are not available a priori.

\subsection{Context Window Scaling and Multi-Chunking Approaches}
Several works attempt to fundamentally extend LMMs' spatial or temporal context windows, either through architectural redesigns targeted at extreme sequence lengths~\citep{chen2025longvila,zhang2025long}, or by using multi-chunked, hierarchical architectures that rely on vast computational resources~\citep{li2025videochatflashhierarchicalcompressionlongcontext}. While such solutions demonstrably push the feasible upper bound of context length, their substantial resource requirements often constrain practical deployment, especially outside large industrial or academic labs. In contrast, our focus is to maximize long-horizon reasoning and accuracy within budgets and token limits compatible with standard academic research or production hardware (see the main text and Appendix~I for further cost analysis).
A similar distinction applies to Video-XL~\citep{videovlcvpr}, which also addresses hour-scale video. While effective, its aggregation mechanism operates \textit{within} the LLM's architecture. Our approach, by contrast, is a fully \textit{pre-LLM} compression module. Furthermore, the temporal selection methods differ significantly: Video-XL employs CLIP-based selection, whereas our TAA uses a learnable, $\Delta_t$-based policy derived from the SSM itself.

\subsection{Mamba-based Compression: Prior Extensions and Our Distinction}
Recent advances have incorporated state-space models (especially Mamba-based architectures) for scalable video modeling in LMMs~\citep{vamba,bimba,jiang2025tokenefficientlongvideounderstanding}. VAMBA~\citep{vamba} integrates a cross-attention structure with Mamba directly within the LLM backbone, with a hybrid junction module that blends visual and textual representations. While this approach enables direct temporal-context association within multimodal tokens, it also entails a significant increase in model parameter size (e.g., Qwen-7B to $\sim$10B), which diverges from our lightweight, modular input compression focus. Furthermore, the point of architectural intervention in VAMBA is fundamentally different from our approach: whereas VAMBA interleaves compression inside the language model, our method applies adaptive, learnable compression externally, as a modular component prior to the LLM input.

BIMBA~\citep{bimba} adopts a different strategy, using periodic, non-parametric anchor queries together with bidirectional Mamba blocks and 3D average pooling to distill salient video frames. In contrast, our approach employs learned, adaptive gating for spatial aggregation and a secondary delta-based temporal selection mechanism, directly addressing feature redundancy and temporal localization bottlenecks as input lengths extend into the hundreds of frames. These design choices, validated through extensive stress-testing (Appendix~J), enable our model to maintain stable compression performance at input scales (e.g., 300--500+ frames) not directly addressed by BIMBA methods.

STORM~\citep{jiang2025tokenefficientlongvideounderstanding} proposes a Mamba-based temporal encoder situated between the visual encoder and LLM, enriching image tokens with temporal information before applying a combination of token compression schemes such as temporal pooling, spatial pooling, and test-time temporal sampling. This architecture delivers strong efficiency/accuracy trade-offs on long-video benchmarks. However, STORM typically employs a mix of several compression strategies simultaneously, blending spatial and temporal pooling with SSM modeling, which can make it challenging to distinguish the effectiveness of individual mechanisms. Furthermore, while STORM effectively reduces the token load to the LLM, its design does not directly leverage fine-grained SSM properties (such as explicit modeling of temporal deltas) as in our method.

\subsection{Empirical Benchmarks, Initialization Protocols, and Our Position}
Across public long video LMM benchmarks~\citep{wang2024lvbench,MLVU,fu2024video,longvideobench,zhao2024videoniah,xiao2021next}, we observe that the empirical performance of recent compression-based approaches is sensitive not only to architectural choices but also to initialization and evaluation protocols. 
Kangaroo~\citep{liu2024kangaroo} achieves strong performance, but its philosophy contrasts fundamentally with ours. The authors attribute its success to a massive-scale curated dataset and a complex curriculum training pipeline, pairing a standard MLP projector with a Llama-3-8B backbone. This contrasts with our focus on a compact data size and a novel modular compression architecture.
STORM~\citep{jiang2025tokenefficientlongvideounderstanding} achieves its high benchmark scores by initializing from Qwen2-VL---a large-scale proprietary vision-language backbone---and then applying additional fine-tuning.
Similarly, a final model in BIMBA~\citep{bimba} reports higher scores (e.g., 59.46 on LongV~\citep{longvideobench}) that result from \textit{further fine-tuning} on an already-aligned LLaVA-Video backbone, which is not a matched comparison to our from-scratch module training.
\textbf{In this work, we intentionally avoid adopting such transfer-based development pipelines and do not consider models initialized from or heavily dependent on proprietary or high-resource pretrained multimodal LLM backbones as direct baselines in our main paper.} We believe this is essential for scientific clarity, as reliance on such initialization pipelines risks amplifying dataset/model bias and undermining generality and fairness in architectural evaluation.

By contrast, our approach deliberately avoids transferring multimodal parameters or prealignments from existing MLLMs (such as Qwen2-VL) and instead emphasizes either from-scratch or strictly open-source initializations for the multimodal architecture. While our models still leverage pretrained language models as the text backbone, \textbf{all multimodal fusion and compression modules are trained from-scratch}, ensuring that our ablation studies purely isolate the effects of our architectural and algorithmic contributions.

Our approach maintains strong competitiveness or state-of-the-art results at globally low token budgets, with scalable performance as input length increases. We also demonstrate the stability of learned and adaptive compression when handling hundreds of frames. In summary, while prior works have substantially advanced LMMs through a range of Mamba-based or pooling-based compression strategies, our approach is distinguished by its modular design, adaptive selection mechanisms, rigorous evaluation methodology, and principled architectural choices.

\section{Model Implementation Details}
\label{app:impl_details}

\paragraph{Overall Architecture}
Our model consists of a multi-stage architecture analogous to LLaVA~\citep{liu2023llava}, but with a dedicated in-network compression flow. Input frames are first processed by a Vision Encoder (SigLIP2-SO-400M~\citep{tschannen2025siglip}), which outputs embeddings of shape \texttt{[batch, frames, 576, 1152]} (where each frame yields $24\times24=576$ patch tokens at $384\times384$ input resolution). 
These embeddings are then fed to a \textbf{Spatiotemporal Compressor Layer} comprising 3 stacked bi-directional Mamba2 blocks, which aggregates patch tokens row-wise into $k$ learned queries per frame. (See Fig.~1 and Fig.~2 in the main text for a schematic overview.)

This output is then passed to a single \textbf{Time Axis Aggregator (TAA) Layer}, which further compacts temporal information. The TAA can be either a uni- or bi-directional Mamba2 layer; however, in our final setting, we use the uni-directional layer for the TAA. The TAA is surrounded by a linear bottleneck adapter (see Table~\ref{tab:hyperparams}), which facilitates reduction and restoration of sequence dimensionality. During training, TAA performs uniform frame subsampling, while during inference, delta-based secondary sampling is applied (see main text and Appendix~K). The resulting representation hence follows the flow:
\begin{itemize}
    \item \texttt{[batch, frames, 576, 1152]} $\rightarrow$ (Compressor) $\rightarrow$ \texttt{[batch, frames, $k$, $d_\mathrm{hidden}$]}
    \item $\rightarrow$ (TAA + sampling) $\rightarrow$ \texttt{[batch, sampled\_frames, $k$, $d_\mathrm{hidden}$]}
    \item $\rightarrow$ (reshape + linear map) $\rightarrow$ \texttt{[batch, sampled\_frames {\footnotesize$\times$} $k$, $d_\mathrm{llm}$]}
\end{itemize}
where the final 1D vision feature sequence is supplied as input tokens to the Qwen2-7B LLM.

The total learnable parameters for the proposed module are $\sim$247M. All parametric baselines compared in Figures~3--4 are scaled to similar parameter counts (246--279M).
Key architectural hyperparameters (e.g., $d_\mathrm{hidden}$, $k$, $d_\mathrm{llm}$) are detailed in Appendix~E.
For further details on module structure and code, refer to Figures~1,~2, and our GitHub repository.

\paragraph{Preprocessing}
We follow the default preprocessing pipeline of LLaVA-OneVision~\citep{li2025llavaonevision}, including video frame extraction, resizing, and normalization. For more details on the data preprocessing, please refer to Appendix~D.

\paragraph{Training Configuration}
Adam~\citep{adam} optimizer with 3\% linear warm-up and cosine annealing is used throughout. Frame sampling is uniform across the video duration, capped at a maximum number of frames ($M$) and FPS (see Appendix~\ref{app:hyperparams}); for longer videos, effective FPS is adaptively reduced within budget. For numerical stability, gate values in the gating mechanism are constrained within $g\in[\epsilon,\,1{-}\epsilon]$ (see Appendix~\ref{app:hyperparams}). Training is performed with bfloat16 precision for memory efficiency and speed. Compressor and aggregator modules are trained with all frames in parallel during training. Delta-based frame selection applies only at inference and is described further in the main text and Appendix~K.

All large-scale experiments utilize PyTorch FSDP~\citep{zhao2023pytorchfsdpexperiencesscaling} with 8 $\times$ A100 GPUs; controlled architecture ablations take $\sim$5.4 hours (i.e., approx 2 A100-days). For the scaled final model training, it requires approx $\sim$45 A100-days.
A comprehensive list of all training hyperparameters is provided in Appendix~E.

\paragraph{Reproducibility Statement}
The code, model weights, and configurations are publicly available at our GitHub repository.

\section{Computing Infrastructure}

All experiments are conducted on a computing server equipped with NVIDIA A100 GPUs (80GB) and Intel Xeon Gold 6338 CPUs, running Ubuntu 22.04.5 LTS. We use PyTorch (version 2.4.1) and CUDA (version 12.2) as the primary deep learning framework and runtime. Additional key software includes the \texttt{transformers} library (v4.50.0) and \texttt{accelerate} (v1.0.1).
Relevant package versions and dependencies for experiment reproducibility are fully documented in the code repository.
Our results are expected to be reproducible on machines with similar hardware and software configurations.

\section{Training Data and Preprocessing}

We provide here a detailed description of the datasets and preprocessing strategies used for multimodal model training in all our experiments. Rather than relying on incremental training of existing multimodal LLMs, we construct our pipeline such that large language and vision components are first aligned on images, followed by joint video training. This enables broad applicability and strict reproducibility. As described in the main paper, we refer to this approach as a ``clean'' and ``from-scratch'' pipeline, emphasizing the transparent and fully controlled nature of our model development. All datasets used for model pre-training and fine-tuning are publicly available, thereby ensuring the full reproducibility of our experimental pipeline.

\paragraph{Module Pre-Training Dataset}
For the lightweight alignment of our proposed compression module, we use the publicly available LLaVA-Pretrain dataset~\citep{liu2023llava}, which contains approximately 558K caption-style image-text pairs.

\paragraph{Image-Instruct Tuning Dataset}
We follow the open multimodal dataset recipe from Elva~\citep{kim-seo-2024-efficient}, providing a diverse set of instructional image datasets. To maintain broad generalization, we exclude samples tailored to highly specialized tasks (e.g., object detection and bounding box annotations) and also omit Vision Flan~\citep{xu-etal-2024-vision} subsets. After these exclusions, our final image-instruction tuning dataset comprises roughly 1 million examples. 

\paragraph{Video-Instruct Tuning Dataset}
For video instruction tuning, we use the large-scale dataset introduced by~\citet{zhang2024video}, providing QA pairs from 178K unique videos. We also utilize a subset of Vista-Set~\citep{ren2024vistaenhancinglongdurationhighresolution}, specifically those related to long video captioning and event relation question answering. This results in a combined training corpus of over 1.4M+ samples. For ablation studies, to ensure feasible compute, we curate a 133K sample subset focused on videos under 30 seconds from~\citet{zhang2024video} and the long video caption portion of~\citet{ren2024vistaenhancinglongdurationhighresolution}.

\paragraph{Preprocessing Details}
To efficiently handle large-scale image and video data, we use the LMDB~\citep{lmdb} for fast disk access. Rather than aggregating all conversational annotations into a single JSON file, we convert the data into JSONL (JSON Lines) format, where each entry is stored on a separate line. This significantly improves memory usage and facilitates efficient streaming of individual samples during training. For video preprocessing, instead of extracting frames on-the-fly during training using libraries such as \texttt{decord}~\citep{decord}, we pre-extract frames in advance based on the known maximum number of frames and frames per second required by each experiment (e.g., 64 frames for ablation studies and 128 frames for main experiments). The extracted frames are stored as multi-image datasets in the LMDB format, enabling faster and more reproducible experimental iterations. This preprocessing strategy is uniformly applied to all baselines and proposed models to ensure fairness and consistency.

\section{Full Hyperparameter Listing and Scripts}
\label{app:hyperparams}

\subsection{Hyperparameter Summary}
We summarize all key hyperparameters for architecture, training, and ablations in Table~\ref{tab:hyperparams}. These parameters correspond to the discussions in Appendix~B and the paper.

\begin{table}[t]
\centering
\begin{adjustbox}{width=\linewidth}
\begin{tabular}{lc}
\toprule
\textbf{Hyperparameter} & \textbf{Value} \\
\midrule
\multicolumn{2}{l}{\textbf{\textit{Architecture}}} \\
\quad Spatiotemporal Compressor blocks & 3 \\
\quad Compressor hidden size ($d_\mathrm{hidden}$) & 3072 \\
\quad Compressor Mamba heads ($n_\mathrm{heads}$) & 96 \\
\quad Compressor Mamba head dim ($d_\mathrm{head}$) & 64 \\
\quad Queries per frame ($k$) & 24 \\
\quad TAA bottleneck adapter & 3072 $\rightarrow$ 128 $\rightarrow$ 3072 \\
\quad LLM embedding dim ($d_\mathrm{llm}$) (Qwen2-7B) & 3584 \\
\quad GPA gate stability ($\epsilon$) & 0.01 \\
\midrule
\multicolumn{2}{l}{\textbf{\textit{Training (General)}}} \\
\quad Optimizer & Adam \\
\quad Adam $\epsilon$ & $1\times10^{-6}$ \\
\quad Max gradient norm & 1.0 \\
\quad Weight decay & None \\
\quad Warmup ratio & 0.03 \\
\quad LR scheduler & Cosine \\
\midrule
\multicolumn{2}{l}{\textbf{\textit{Module Pre-Training}}} \\
\quad Learning rate & $1\times10^{-4}$ \\
\quad Batch size & 768 \\
\quad Steps & 727 \\
\quad Epochs & 1 \\
\midrule
\multicolumn{2}{l}{\textbf{\textit{Image-Instruct Tuning}}} \\
\quad Learning rate & $2\times10^{-5}$ \\
\quad Batch size & 128 \\
\quad Steps & 8K \\
\midrule
\multicolumn{2}{l}{\textbf{\textit{Video-Instruct Tuning (Main)}}} \\
\quad Learning rate & $2\times10^{-5}$ \\
\quad Max input frames ($M$) & 128 \\
\quad Max input FPS & 2.0 \\
\quad Batch size & 768 \\
\quad Steps & 2K \\
\midrule
\multicolumn{2}{l}{\textbf{\textit{Video-Instruct Tuning (Ablation)}}} \\
\quad Learning rate & $2\times10^{-5}$ \\
\quad Max input frames ($M$) & 64 \\
\quad Max input FPS & 2.0 \\
\quad Batch size & 72 \\
\quad Steps & 2K \\
\bottomrule
\end{tabular}
\end{adjustbox}
\caption{Comprehensive summary of key architectural and training hyperparameters.}
\label{tab:hyperparams}
\end{table}

\subsection{Training}
We provide concise summaries of the training script configurations used in our study for three major stages: module pre-training, image-instruct tuning, and video-instruct tuning. All key hyperparameters are listed in Table~\ref{tab:hyperparams}.

The training process follows standard LLaVA-style recipes.
All experiments use the Adam optimizer with cosine learning rate scheduling and a warmup ratio of 0.03.
All scripts leverage the \texttt{accelerate} launcher for distributed and mixed-precision training. 
The launch commands, configs, and scripts are publicly available in our code repository.

\subsection{Inference}
All model evaluations are conducted using the \texttt{lmms-eval}~\citep{zhang2024lmmsevalrealitycheckevaluation} library in a standardized environment. Unless noted otherwise, the following inference settings apply to the experiments:
\begin{itemize}
    \item For ablation experiments (e.g., Table~2), models trained with 64-frame inputs are evaluated with 256 frames at inference.
    \item For the final model results (Table~1 and Table~3), models trained with 128-frame inputs and evaluated with 384 frames at inference.
    \item The experiments with Pythia~\citep{biderman2023pythiasuiteanalyzinglarge} and Mamba2~\citep{mamba2} are conducted using both training and inference with 128-frame inputs (see Table~4).
    \item Any exceptions to the above (such as for Figure~3 and Figure~4) are explicitly indicated in corresponding captions or the main text.
    \item All results are computed using greedy decoding.
\end{itemize}
These settings ensure comparability across benchmarks and facilitate clear interpretation of the experimental results.

\section{Evaluation Benchmarks}

This section introduces the set of video understanding benchmarks used in our evaluation. Each benchmark covers distinct aspects of video understanding, providing a rigorous ground for assessing model performance across tasks such as temporal grounding, fine-grained reasoning, summarization, and causal inference. For reproducibility and precise comparison, we specify the dataset splits and settings employed in our experiments.

\paragraph{LVBench}
LVBench (An Extreme Long Video Understanding Benchmark) is a dedicated benchmark for comprehensive long video understanding, targeting videos that significantly exceed the length of existing datasets. LVBench consists of richly-annotated, open-domain videos drawn from diverse categories. The benchmark evaluates six core capabilities crucial for long video reasoning: Temporal Grounding, Summarization, Reasoning, Entity Recognition, Event Understanding, and Key Information Retrieval. Questions are carefully designed to require multi-step, context-aware analysis, and a rigorous filtering process ensures that visual content must be referenced for correct answers. This benchmark aims to assess long-term temporal memory and comprehension in multimodal models~\citep{wang2024lvbench}.  
For our evaluation, we utilized the official test split of LVBench.

\paragraph{MLVU}
MLVU (Multi-task Long Video Understanding Benchmark) offers an in-depth evaluation framework for long-video understanding across varied genres. With an emphasis on diverse tasks, this benchmark evaluates the multimodal LLMs' proficiency in long video scenarios, uncovering significant areas for improvement. By encompassing extended video lengths and versatile video types, MLVU provides critical insights into the performance of current models and sets the stage for future advancements in long video understanding~\citep{MLVU}.  
All results reported are based on the dev split of the benchmark.

\paragraph{HourVideo}
HourVideo is a benchmark dataset for hour-long video-language understanding, consisting of manually curated egocentric videos from the Ego4D dataset. The benchmark features high-quality, five-way multiple-choice questions across a comprehensive task suite including summarization, perception (recall, tracking), visual reasoning (spatial, temporal, predictive, causal, counterfactual), and navigation (room-to-room, object retrieval). HourVideo is designed to evaluate models' ability to perform fine-grained reasoning over extended temporal contexts~\citep{chandrasegaran2024hourvideo}.
We evaluated on the dev split of the benchmark.

\paragraph{VideoMME}
VideoMME is an evaluation benchmark for multimodal LLMs in video analysis, offering a wide spectrum of video types and temporal durations. This benchmark provides rigorous manual annotations to assess models comprehensively. VideoMME challenges models with a diverse set of visual domains and duration scopes, playing a crucial role in evaluating and understanding the capabilities of MLLMs in handling sequential visual data effectively~\citep{fu2024video}.  
We evaluated on the test split of VideoMME, using the version without subtitle information.

\paragraph{LongVideoBench}
LongVideoBench is a large-scale benchmark designed to evaluate long-form video-language understanding in multimodal models. It consists of 3,763 web videos (up to one hour each) with subtitles and 6,678 human-annotated multiple-choice questions across 17 categories. The core task, referring reasoning, challenges models to retrieve and reason over specific video contexts referenced in each question. By emphasizing detailed, long-context multimodal reasoning, LongVideoBench highlights persistent challenges for both proprietary and open-source LMMs, and serves as a rigorous metric for future advancements in long video-language understanding~\citep{longvideobench}.  
We conducted all evaluations on the official validation split.

\paragraph{VNBench}
VNBench, part of the VideoNIAH framework~\citep{zhao2024videoniah}, provides a unique approach to benchmarking video understanding by inserting synthetic `needles' into videos. It evaluates proficiency in retrieving, ordering, and counting tasks, assessing a model's fine-grained temporal reasoning and dense-video processing capabilities. VNBench adopts a circular evaluation strategy as its official protocol: each query is presented four times with shuffled answer options, and is only considered correct if all four responses are correct, thereby mitigating the effect of random guessing on reported accuracy. In all our experiments, we employ this official circular evaluation strategy. This allows us to fairly and robustly compare our models' fine-grained temporal reasoning abilities under the same rigorous standard.  
All results are computed on the test split of VNBench, following the official evaluation protocol.

\paragraph{NExTQA}
NExTQA is a video QA benchmark focused on explaining video contents, emphasizing causal and temporal reasoning within video interactions. This benchmark aims to guide the development of models that go beyond superficial video understanding towards a more profound temporal and causal comprehension~\citep{xiao2021next}.  
Our experiments use the test split in the multi-choice setting.

\section{Sanity Checks and Baseline Implementations}

To ensure the reliability and reproducibility of our experimental results, we conduct extensive sanity checks using both officially released checkpoints and our own faithful re-implementations of key baseline methods.

\subsection{Open-Model Sanity Check}
We independently evaluate two standard publicly available models—LLaVA-Video-7B-Qwen~\citep{zhang2024video} and LLaVA-OneVision-7B~\citep{li2025llavaonevision}—using their official checkpoints and our experimental setup. Table~\ref{table:sanity} provides a direct comparison between our reproduced results (\(^{\dagger}\)) and the scores reported in the original papers or benchmark leaderboards (\(^{\ddagger}\)). Our results closely track the official numbers, verifying the validity and transparency of our evaluation protocol. For LLaVA-Video, while the original study reports results at 64 input frames (using a higher token count), we standardize our evaluation to 6.3K tokens for fairer and more consistent comparison with our method and other baselines. Across settings, our reproduced performance remains well aligned with published results, further supporting the robustness of our experimental design.

\subsection{Direct Baseline Re-Implementations}
Beyond open-source checkpoints, we also re-implement several key baseline methods based on publicly available code and paper descriptions. For the average pooling baseline, we closely follow the LLaVA-OneVision codebase, as the methodology is clearly specified and straightforward to reproduce.

For the BIMBA baseline—which is particularly relevant for comparison with our method—we reference both the original paper and the authors' released code. In our implementation, we strictly adhere to the methodological details provided in the manuscript, aiming to faithfully reflect their methodological framework (rather than reproducing a specific variant checkpoint). Specifically, we implement 3D pooling for query token calculation, matching the kernel sizes and configurations as described by the authors: 64 frames with $24 \times 24$ patch token features are compressed into $16 \times 12 \times 12$ tokens via a 3D average pooling operation. Although minor differences in layer counts or hyperparameters may exist, we ensured that all critical architectural and compression settings followed the original BIMBA methodology.

To enable a direct comparison, we evaluate our BIMBA implementation under the same conditions as reported in \citet{bimba} (using Vicuna-7B, 64 input frames, and equivalent compression settings). Our main results utilize a training set of 133K QA samples (Mini setting), which is smaller than the original BIMBA training data (approximately 370K samples) but sufficient to reveal overall trends. The results are summarized in Table~\ref{table:bimba-compare}, which compares (i) the original BIMBA results, (ii) our faithful baseline implementation (BIMBA$^\prime$), and (iii) our proposed model under identical data and evaluation conditions. Our BIMBA baseline matches or exceeds the original scores, confirming the fidelity of our implementation. Most importantly, our proposed model consistently outperforms both the original and our BIMBA-based baselines by a notable margin.

\begin{table}[t!]
\centering
\begin{adjustbox}{max width=\linewidth}
\begin{tabular}{lccc}
    \toprule
    \textbf{Model} (Data Size)        & \textbf{MLVU} & \textbf{MME} & \textbf{NQA} \\
    \midrule
    BIMBA (reported in~\citet{bimba}, 370K)   & 47.2 & 45.7 & 72.4 \\
    BIMBA$^\prime$ (our implementation, 133K)    & 55.5 & 44.8 & 72.7 \\
    \rowcolor{green!20}
    Proposed Mini (133K) & \textbf{58.5} & \textbf{47.8} & \textbf{73.3} \\
    \bottomrule
\end{tabular}
\end{adjustbox}
\caption{\textbf{Comparison of BIMBA and our proposed method under matched conditions.} Our BIMBA implementation closely replicates the architectural details of the original (including 3D pooling: $64 \rightarrow 16 \times 12 \times 12$), and achieves similar or higher scores on key benchmarks. Our proposed model further improves upon these baselines in all respects.}
\label{table:bimba-compare}
\end{table}

These sanity checks and carefully designed baseline comparisons demonstrate the overall fairness, trustworthiness, and reproducibility of our experimental setup. They also highlight the improvements delivered by our proposed method over prior baselines.

\begin{table}[t!]
\centering
\begin{adjustbox}{width=\linewidth}
\begin{tabular}{lcccccc}
    \toprule
    \textbf{Model} & \textbf{\#Frame / \#Tok} & \textbf{MLVU} & \textbf{MME} & \textbf{LongV} & \textbf{NQA} & \textbf{Avg.} \\
    \midrule
    LLaVA-OneVision$^{\dagger}$ & 32 / 6,272 &  65.2 & 58.5 & 56.6 & 79.3 & 64.9\\
    LLaVA-OneVision$^{\ddagger}$ & 32 / 6,272 & 64.7 & 58.2 & 56.5 & 79.4 & 64.7 \\
    \midrule
    LLaVA-Video$^{\dagger}$ & 32 / 6,272 &  66.3 & 62.4 & 58.0 & 81.8 & 67.1\\
    LLaVA-Video$^{\ddagger}$ & 64 / 12,544 & 70.8 & 63.3 & 58.2 & 83.2 & 68.9 \\
    \bottomrule
\end{tabular}
\end{adjustbox}
\caption{\textbf{Baseline Performance Check.}  
Score comparison between previously reported results ($^{\ddagger}$; taken from their original papers, recent literature, and official benchmark leaderboards) and our reproduced evaluation results ($^{\dagger}$).}  
\label{table:sanity}
\end{table}

\section{Randomness Analyses and Statistical Tests}
\label{sec:randomness}

To assess the reliability and reproducibility of our results, all primary experiments in this work use a fixed random seed (``\texttt{2025}''), unless otherwise specified. In this section, where computationally practical, we conduct multiple independent training trials to quantify run-to-run variability and assess the statistical significance of key findings. Below, we summarize both the variance observed across trials and the robustness of our conclusions.

\subsection{Repeatability of Final Model Configurations.}
For the final large-scale experiments, Table~\ref{tab:reproducibility-qwen} presents results from two fully independent training runs, each using a distinct random seed. For the Proposed Mini Model (Vicuna-7B), Table~\ref{tab:reproducibility-mini} reports results from four independent trials. The consistently low standard deviations across all metrics indicate stable and reproducible convergence. Importantly, the overall trends are unchanged under different random seeds, supporting the robustness of our experimental conclusions.

\begin{table}[t]
    \centering
    \begin{adjustbox}{width=\linewidth}
    \begin{tabular}{lcccc}
        \toprule
        & LVBench & MLVU & MME & Avg. \\
        \midrule
        Exp. 1 & 44.61 & 67.99 & 58.26 & 56.95 \\
        Exp. 2 & 44.93 & 67.50 & 58.56 & 57.00 \\
        \midrule
        \textbf{Mean (Std)} & 44.77 (0.16) & 67.75 (0.24) & 58.41 (0.15) & 56.98 (0.02) \\
        \bottomrule
    \end{tabular}
    \end{adjustbox}
    \caption{
        \textbf{Repeatability of Proposed Model (Qwen2-7B).}
        Results from two fully independent training trials. Each run uses a distinct random seed, showing consistent performance.
    }
    \label{tab:reproducibility-qwen}
\end{table}

\begin{table}[t]
    \centering
    \begin{adjustbox}{width=\linewidth}
    \begin{tabular}{lcccc}
        \toprule
        & LVBench & MLVU & MME & Avg. \\
        \midrule
        Exp. 1 & 37.90 & 58.50 & 47.80 & 48.07 \\
        Exp. 2 & 35.80 & 56.90 & 48.20 & 46.97 \\
        Exp. 3 & 36.20 & 58.50 & 46.10 & 46.93 \\
        Exp. 4 & 36.30 & 58.00 & 47.60 & 47.30 \\
        \midrule
        \textbf{Mean (Std)} & 36.55 (0.80) & 57.98 (0.65) & 47.42 (0.79) & 47.32 (0.46) \\
        \bottomrule
    \end{tabular}
    \end{adjustbox}
    \caption{
        \textbf{Repeatability of Mini Model (Vicuna-7B).}
        Performance over four fully independent runs, highlighting stable and reproducible convergence.
    }
    \label{tab:reproducibility-mini}
\end{table}

\begin{table}[t]
    \centering
    \begin{adjustbox}{width=\linewidth}
    \begin{tabular}{lcccc}
        \toprule
        & LVBench & MLVU & MME & Avg. \\
        \midrule
        No GPA, Averaged Query & 34.67 (0.48) & 53.18 (0.67) & 47.58 (0.23) & 45.14 (0.24) \\
        \rowcolor{green!20}
        With GPA     & 41.10 (0.40) & 63.01 (0.59) & 53.82 (1.02) & 52.64 (0.43) \\
        \bottomrule
    \end{tabular}
    \end{adjustbox}
    \caption{
        \textbf{Ablation of GPA Module.}
        Mean and standard deviation over three independent trials per configuration. GPA consistently improves all benchmarks.
    }
    \label{tab:gpa-ablation}
\end{table}

\begin{table}[t]
    \centering
    \begin{adjustbox}{width=\linewidth}
    \begin{tabular}{lcccccc}
        \toprule
        & 1 & 2 & 3 & 4 & 5 & 6 \\
        \midrule
        Uniform  & 56.30 & 56.67 & \textbf{48.33} & 46.65 & 46.74 & 46.66 \\
        \rowcolor{green!20}
        Delta-Based  & \textbf{56.95} & \textbf{57.00 }& 48.07 & \textbf{46.97} & \textbf{46.93} & \textbf{47.30} \\
        \bottomrule
    \end{tabular}
    \end{adjustbox}
    \caption{
        \textbf{Delta-based vs. Uniform Sampling (Wilcoxon Test).}
        Six fully independent trials used as statistical units. Delta-based sampling yields consistently higher scores.
    }
    \label{tab:wilcoxon}
\end{table}

\subsection{Robustness of the GPA Module}
Table~\ref{tab:gpa-ablation} summarizes three independent training runs for each ablation setting. All values are reported as the mean and standard deviation across these independent trials. The GPA module consistently improves performance across all metrics, and the low standard deviations confirm that these gains are robust to random fluctuations.

\subsection{Statistical Analysis of Sampling Methods}
To compare delta-based sampling (our proposal) and uniform sampling, we aggregate six fully independent training trials—two for Qwen2-7B and four for the Vicuna-based Mini model. Although these trials originate from two backbones, each trial involves a complete, independent training process using a unique random seed and data shuffling. This approach ensures that each score represents a valid, statistically independent experimental unit. Importantly, for the purpose of evaluating the general effectiveness of the sampling method itself, we treat all six trials equally and focus solely on the relative impact of delta-based versus uniform sampling. This avoids pseudo-replication and ensures that our statistical comparison is both fair and interpretable. Applying the paired Wilcoxon signed-rank test (\texttt{scipy.stats.wilcoxon}) yields a test statistic of 2.000 and a p-value of 0.04688, indicating that the improvement achieved by delta-based sampling is statistically significant at the conventional $\alpha = 0.05$ level.

Across all main experiments and ablation studies, our results consistently exhibit a high degree of reproducibility under independent stochastic trials. Furthermore, statistical testing supports the reliability of the observed trends. While some variability is inevitable in stochastic training, these analyses suggest that the principal improvements reported in this work are robust to random seeds and sampling effects. 
Our code is publicly available to facilitate transparency and reproducibility.
We hope that this work will contribute to progress in the field and serve as a helpful reference for future research.

\section{Inference and Training Cost Analysis}

In line with the primary objective of our study—hour-long video understanding—all measurements of latency (throughput), memory consumption, and maximum LLM token usage (i.e., vision token count) are performed on LVBench~\citep{wang2024lvbench}. We report the first-token latency, which specifically captures the delay incurred in generating the initial output token. This effectively reflects the resource cost during the prefill stage of inference, which is particularly critical for hour-long video analysis.

The measurement setup is identical to that described in Appendix~C: all experiments are conducted on a single NVIDIA A100 80GB GPU using FP16 inference. To ensure reliability, the first 50 inference runs serve as warm-up. The subsequent 200 inference samples are used to compute the average inference speed and memory usage. Performance metrics measured under these standardized conditions are summarized and visualized to provide a comparative analysis of model efficiency for large-scale video understanding tasks.

\section{Further Limit Testing}

To better understand the robustness and latent boundaries of our model, we conduct two sets of stress tests that go beyond the standard settings in the main paper. The first set systematically varies the threshold parameter $\delta_{\text{thresh}}$ used in delta-based sampling, measuring effects on overall performance, compression ratio, and effective vision token count. The second set expands the number of input frames far beyond typical scales to assess the impact on LVBench performance.

\paragraph{Ablation on Delta-Time Threshold ($\delta_{\text{thresh}}$)}

In the main experiments, we primarily set $\delta_{\text{thresh}} = 0.6$ for secondary sampling to strike a balance between information retention and sampling efficiency. To examine the robustness of our approach, we further conduct limit tests by systematically lowering or increasing this threshold. Increasing $\delta_{\text{thresh}}$ aggressively filters out more frames, yielding higher compression rates but often at the expense of end-task accuracy. In contrast, decreasing $\delta_{\text{thresh}}$ preserves more frames, lowering the compression rate and resulting in a substantial increase in vision token usage.
Table~\ref{table:thres} summarizes the LVBench performance, the mean sampling rate, and the maximum number of vision tokens passed to the language model under varying threshold settings.

\begin{table}[t!]
\centering
\begin{adjustbox}{width=0.95\linewidth}
\begin{tabular}{cccc}
    \toprule
    $\delta_{\text{thresh}}$ & \textbf{LVBench} & \textbf{Sampling Rate} & \textbf{Max Vision Token} \\
    \midrule
    0.2 & 45.58 & 100.0 & 9,216 \\
    0.4 & 44.67 & 75.24 & 8,856 \\
    \rowcolor{green!20}
    0.6 & 44.61 & 51.04 & 4,704 \\
    0.8 & 43.45 & 44.22 & 4,584 \\
    1.0 & 43.32 & 33.74 & 3,504 \\
    1.2 & 42.09 & 30.42 & 3,072 \\
    \bottomrule
\end{tabular}
\end{adjustbox}
\caption{\textbf{Delta-Time Threshold Ablation on LVBench.} As the threshold increases, more aggressive frame selection leads to increased compression (lower sampling rate and vision token count), but at the cost of moderate performance drop. Notably, model degradation is gradual rather than catastrophic even at extreme compression, underscoring the algorithmic robustness of our temporal gating module.}
\label{table:thres}
\end{table}

\paragraph{Scaling Input Frame Length}
We next evaluate our final model's performance while scaling the number of input frames well beyond the default values. Table~\ref{table:frame-scale} presents the LVBench score as we increase the input length. We observe clear gains up to 384 input frames, after which the performance plateaus, indicating efficient long-context utilization but no further benefits beyond a certain point.
These limit tests demonstrate that our model maintains stability and practical utility across a broad range of compression settings and input scales, with no catastrophic failures or degenerate behaviors under heavy stress. This highlights the resilience and efficiency of our approach for long-context, resource-constrained video understanding tasks.

\begin{table}[t!]
\centering
\begin{adjustbox}{max width=0.5\linewidth}
\begin{tabular}{cc}
    \toprule
    \textbf{Input Frames} & \textbf{LVBench} \\
    \midrule
    128  & 43.58 \\
    256  & 44.87 \\
    \rowcolor{green!20}
    384  & 45.58 \\
    512  & 45.45 \\
    \bottomrule
\end{tabular}
\end{adjustbox}
\caption{\textbf{LVBench Performance vs. Input Frame Count.} Performance steadily improves up to 384 frames, after which it saturates, suggesting diminishing returns from excessive input length.}
\label{table:frame-scale}
\end{table}

\section{Additional Visualizations and Examples}

We provide further visualizations and qualitative examples to supplement the quantitative results in the main paper, offering additional insight into the behaviors of our modules (GPA and TAA).

\begin{figure}[t!]
     \centering
     \includegraphics[width=0.75\linewidth]{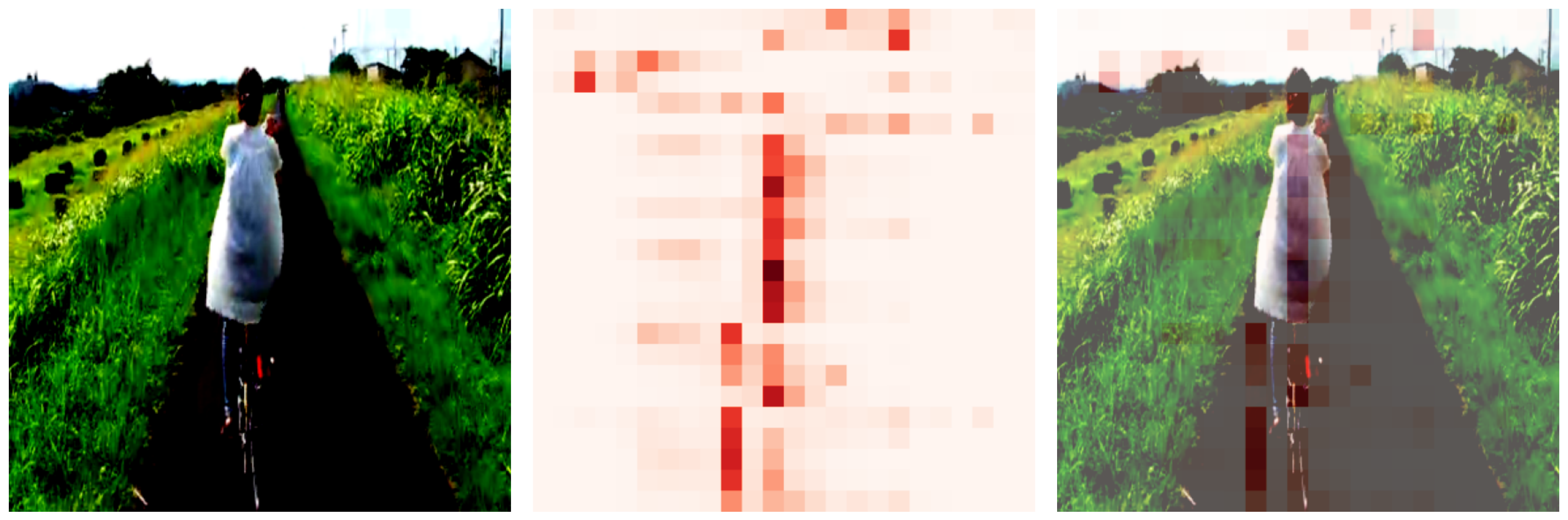}
     \includegraphics[width=0.75\linewidth]{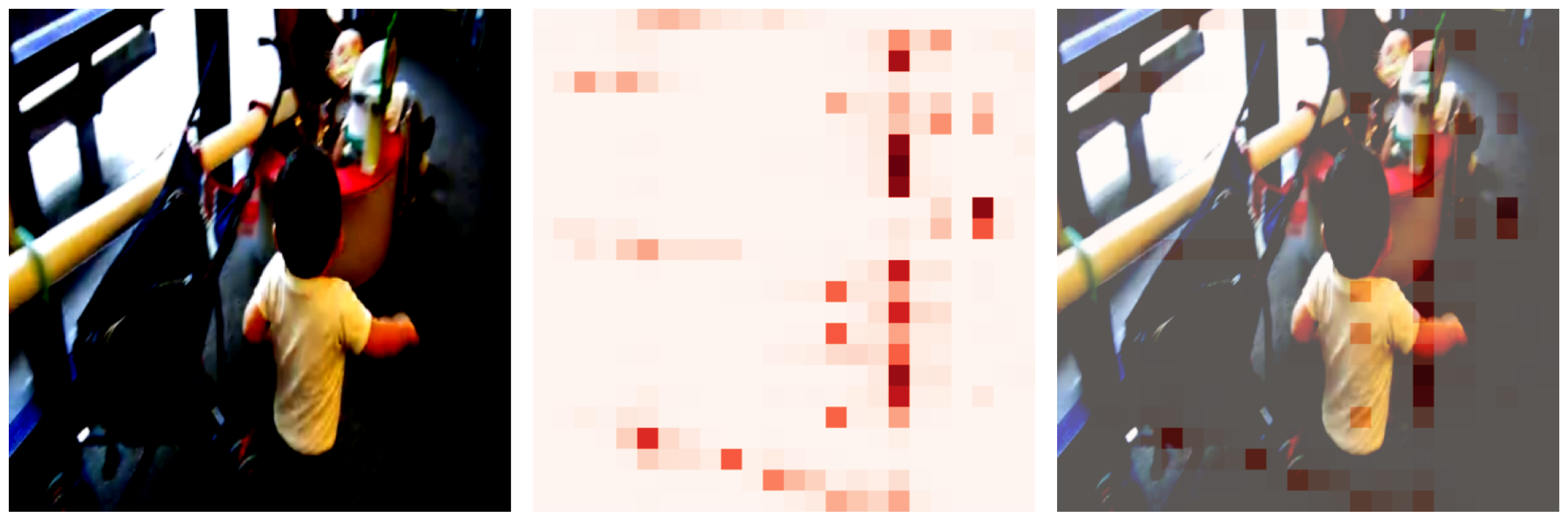}
     \includegraphics[width=0.75\linewidth]{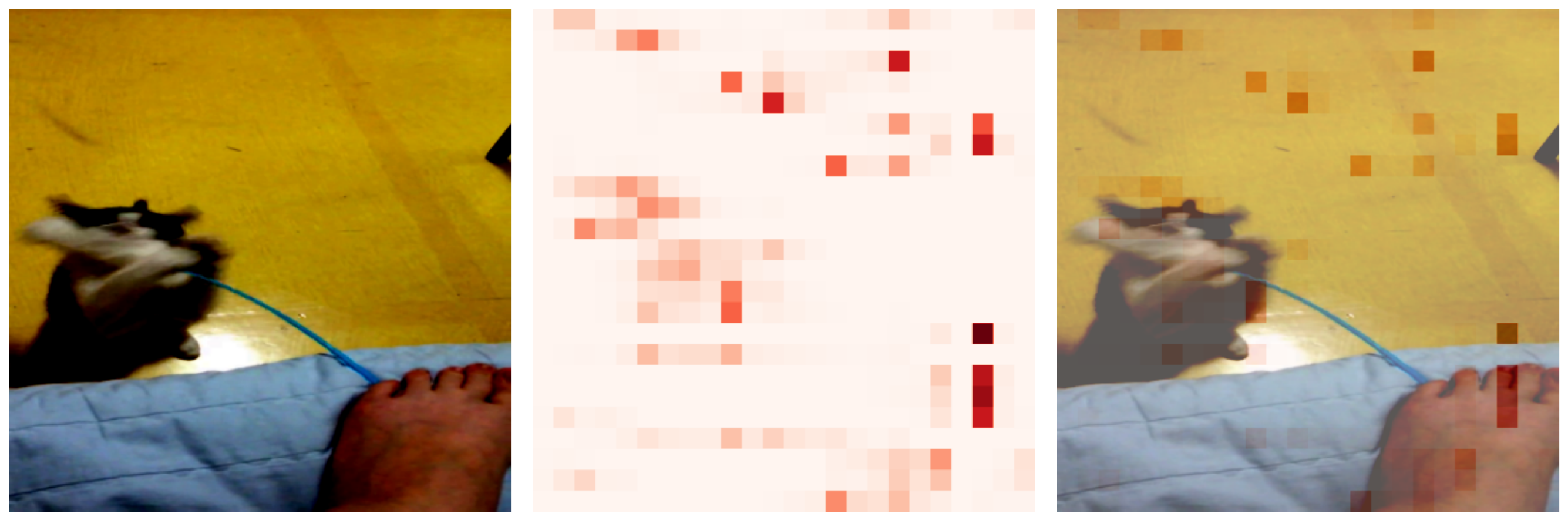}
     \includegraphics[width=0.75\linewidth]{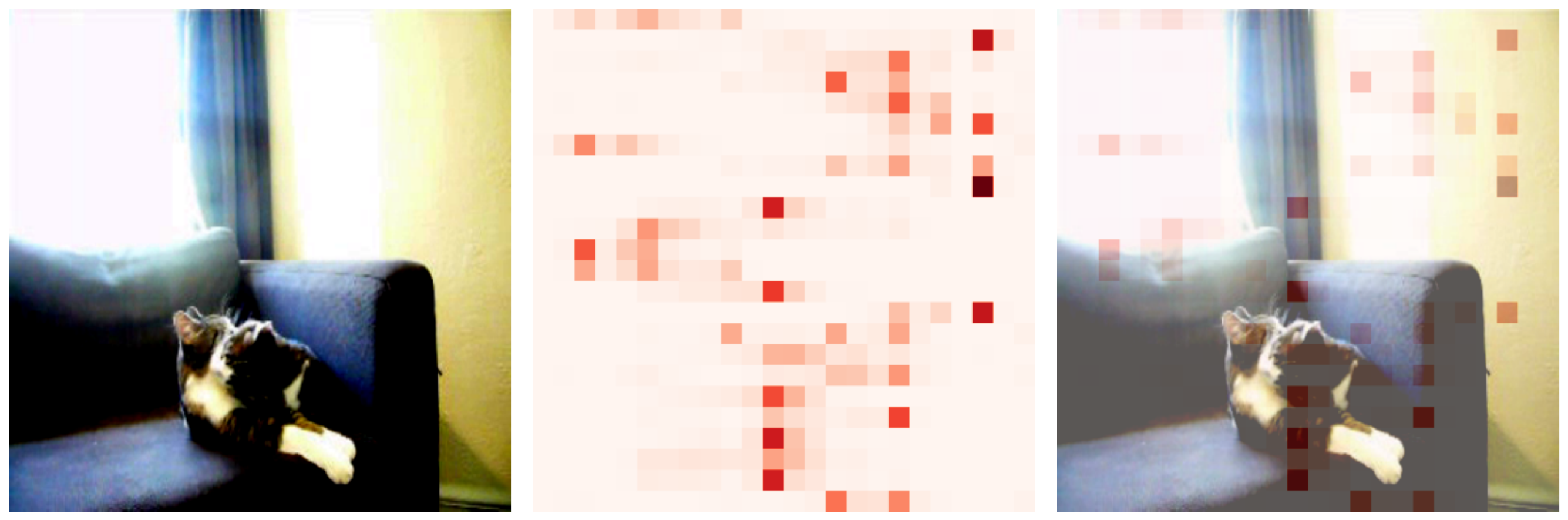}
     \caption{
     Qualitative GPA attention map visualizations on NExT-QA~\cite{xiao2021next} video frames. Across diverse scenes, the module consistently attends to the primary subjects such as people and animals, demonstrating its interpretability and responsiveness to semantically important regions. The attention maps are not pixel-precise but clearly concentrate around key objects in each frame.
     }
     \label{fig:aaai26_gpa_example}
\end{figure}

\begin{figure}[t!]
     \centering
     \includegraphics[width=\linewidth,height=9cm]{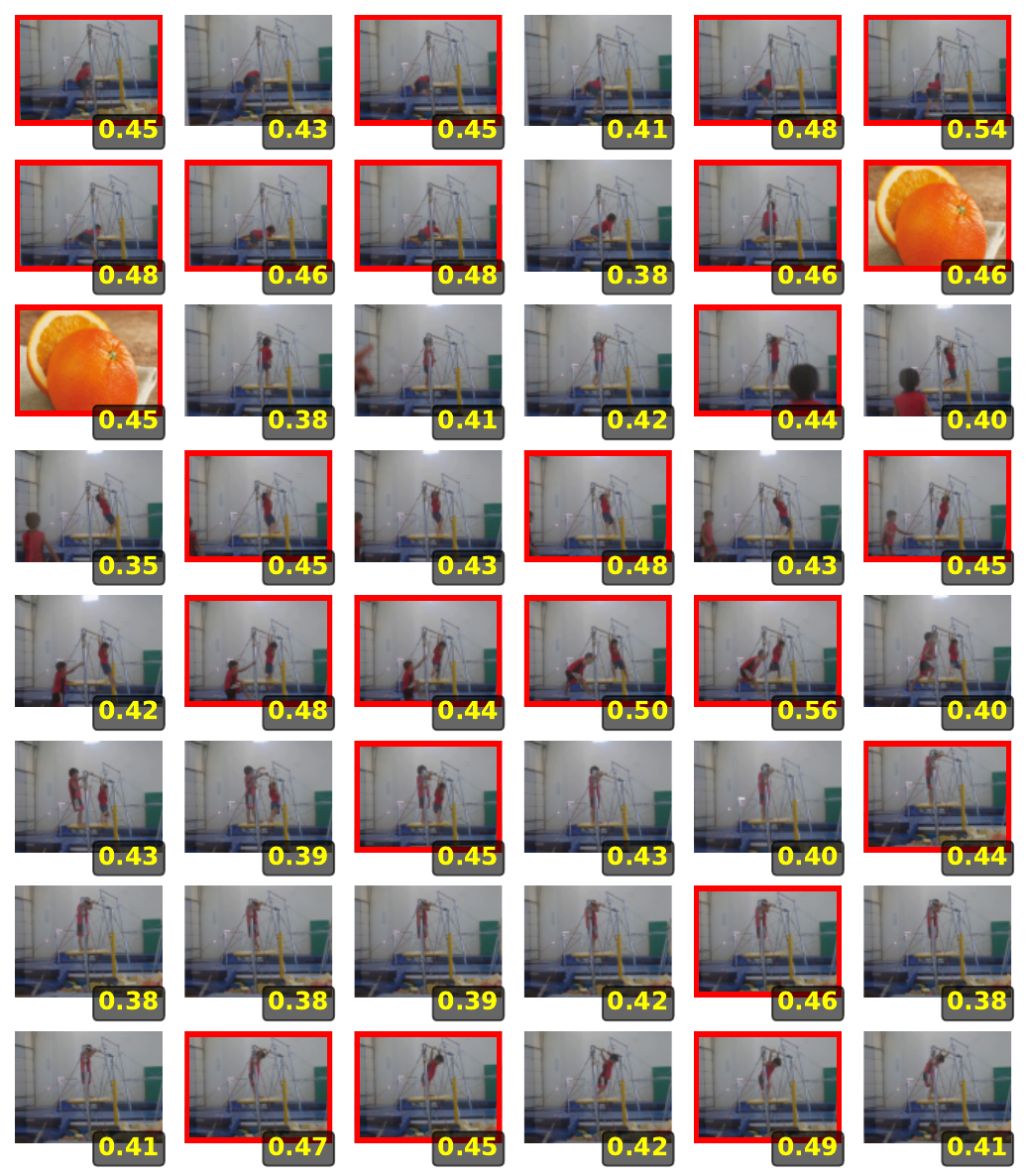}
     \caption{
     Per-frame delta-time values from the TAA layer on a VNbench~\cite{zhao2025needle} sample. Relatively higher values are observed around the embedded needle frame (marked in orange), suggesting that the module does not entirely miss the inserted event, though the overall signal is subtle.
     }
     \label{fig:aaai26_delta_example2}
\end{figure}

\paragraph{GPA Visualization.}
Figure~\ref{fig:aaai26_gpa_example} presents attention maps produced by the GPA module on video frames sampled from the NExT-QA~\cite{xiao2021next} dataset.
Across varied scenes---outdoor activities, indoor settings, and interactions between people and animals---the GPA module consistently highlights regions corresponding to the principal subjects in the frame.
For example, the attention concentrates around persons or a resting cat, indicating that the module is responsive to semantically salient objects.
While the attention maps do not yield pixel-level precision akin to segmentation masks, they clearly localize the key entities, supporting the interpretability of the learned representations.

\paragraph{Per-Frame Delta-Time Value Visualization.}
Figure~\ref{fig:aaai26_delta_example2} visualizes per-frame delta-time values generated by the TAA layer on a video from VNbench~\cite{zhao2025needle}, a needle-in-a-video-haystack benchmark.
While the overall temporal signal is admittedly subtle and not immediately easy to interpret visually, we observe that the delta-time values around the embedded needle frame (the inserted atypical event) tend to be relatively higher compared to the surrounding frames.
This is consistent with the strong quantitative performance on VNbench reported in the main paper: the TAA module does not entirely overlook the needle event even in a qualitative sense.
That said, we acknowledge that the visual contrast is modest, and the primary evidence for the TAA's effectiveness remains the quantitative results rather than these visualizations alone.

\section{Code and Resource Availability}

All source code required to reproduce our results is publicly available at \url{https://github.com/naver-ai/mambamia}.
The repository includes training and inference scripts, configuration files, and instructions to facilitate replication.
Core hyperparameters and experiment settings are listed in Appendix~E.
All datasets used for training and evaluation are also publicly available; dataset references and download instructions are provided in Appendix~D.
Model checkpoints are available via the same repository.

\section{Background on Classical State-Space Models}
\label{sup:classical_ssm}

In control theory literature, a classical continuous-time linear State-Space Model (SSM) represents how input signals affect a latent state which evolves through time. The standard continuous-time formulation is expressed as follows:
\[
h'(t) = \mathbf{A}h(t) + \mathbf{B} x(t),\quad y(t) = \mathbf{C} h(t),
\]
Here, the (latent) hidden state vector \(h(t)\) captures the internal system dynamics. To apply SSMs practically to discrete sequential input (such as video frames or language tokens), we typically adopt a zero-order hold discretization:
\begin{equation*}
\overline{\mathbf{A}}=\exp(\Delta\mathbf{A}),\quad 
\overline{\mathbf{B}}=(\Delta\mathbf{A})^{-1}\bigl(\exp(\Delta\mathbf{A}) -\mathbf{I}\bigr)\Delta\mathbf{B},
\end{equation*}
yielding the discrete form:
\begin{equation*}
h_k=\bar{\mathbf{A}}h_{k-1}+\bar{\mathbf{B}}x_k,\quad y_k=\mathbf{C}h_k,
\end{equation*}
where step-size \(\Delta\) is learnable~\citep{gu2024mamba}. Such discretized recurrence computation clearly results in linear complexity (\(\mathcal{O}(T)\)), efficiently scaling to very large input sequences without the substantial computational overhead associated with Transformer attention models (\(\mathcal{O}(T^2)\)). However, as noted by \citet{gu2024mamba,mamba2}, classical SSMs treat all inputs identically across time-steps, potentially limiting expressive capacity in capturing variety and complexity of inputs, motivating subsequent developments in adaptive (selective) models.

\end{document}